\documentclass[10pt,twocolumn,letterpaper]{article}

\usepackage{cvpr}              

\usepackage{multirow}
\usepackage{ifthen}
\usepackage{graphicx}
\usepackage{algorithm}
\usepackage{comment}
\usepackage{algpseudocode}
\usepackage[table]{xcolor}

\usepackage{xcolor}
\usepackage{amsmath}
\usepackage{amssymb}
\usepackage[table]{xcolor}
\usepackage{threeparttable} 
\usepackage[normalem]{ulem}
\usepackage{wrapfig}
\usepackage{adjustbox}
\usepackage[table]{xcolor}

\usepackage{xcolor}

\newboolean{K_is_15}

\setboolean{K_is_15}{false}

\definecolor{fftcolor}{RGB}{210, 10, 0}     
\definecolor{loracolor}{RGB}{0, 80, 180}    

\definecolor{cvprblue}{rgb}{0.21,0.49,0.74}
\usepackage[pagebackref,breaklinks,colorlinks,allcolors=cvprblue]{hyperref}

\def\confName{CVPR}
\def\confYear{2027}

\title{TACO: Trajectory-Aware Curvature Observation for Memorization Localization and Mitigation in Diffusion Models}

\author{
Ruxi Gu$^{1,2}$ \quad Zilei Wang$^{1}$\textsuperscript{\raisebox{-0.48ex}{\hspace{-0.1em}*}} \quad Wei Wang$^{2,3}$\thanks{Corresponding authors. Emails: zlwang@ustc.edu.cn, wangwei@ustb.edu.cn}\\
\small $^{1}$Department of Automation, University of Science and Technology of China\\[-3pt]
\small $^{2}$State Key Laboratory of General Artificial Intelligence, BIGAI\\[-3pt]
\small $^{3}$School of Computer and Communication Engineering, University of Science and Technology Beijing\\[-3pt]
}

\begin{document}
\maketitle
\begin{abstract}
Text-to-image diffusion models can reproduce memorized training content within localized image regions, making precise localization essential for targeted mitigation. However, curvature observations at a single denoising state can contain transient responses, while spatially uniform mitigation may disrupt prompt-relevant content. We present Trajectory-Aware Curvature Observation (TACO), a unified framework for local memorization detection and mitigation in diffusion models. Our empirical analysis reveals that memorization-related curvature becomes pronounced during late-stage denoising and persists at consistent spatial locations across neighboring states. Motivated by these observations, TACO aggregates prompt-dependent curvature responses along the late denoising trajectory and applies latent-guided spatial refinement to improve spatial coherence. The resulting memorization map serves both as a localization signal and as a spatial control signal for mitigation. Specifically, TACO uses this map to steer existing inference-time interventions toward memorization-related regions through spatial reweighting and localized guidance correction. Extensive experiments demonstrate that TACO consistently outperforms competing localization methods and improves the Pareto trade-off between memorization suppression and generation quality. These results establish trajectory-aware spatial analysis as an effective link between memorization localization and targeted mitigation.
\end{abstract}   

\section{Introduction}
\label{sec:intro}

Text-to-image (T2I) diffusion models \cite{ho2020denoising,song2021scorebased,songdenoising} can reproduce training images or distinctive visual patterns under specific prompts~\cite{carlini2023extracting,somepalli2023diffusion,webster2023reproducible}. Such memorization raises privacy and copyright concerns, as generated outputs may expose sensitive training content or reproduce protected material. Detecting and mitigating this behavior ~\cite{gandikota2023erasing,fan2024salun,park2024direct,alberti2025data,chavhan2024memorized,hintersdorf2024finding,di2026demystifying} is therefore important for deploying T2I models while preserving their ability to generate diverse, prompt-aligned images.

\begin{figure}[t]
    \centering
    \includegraphics[width=\linewidth]{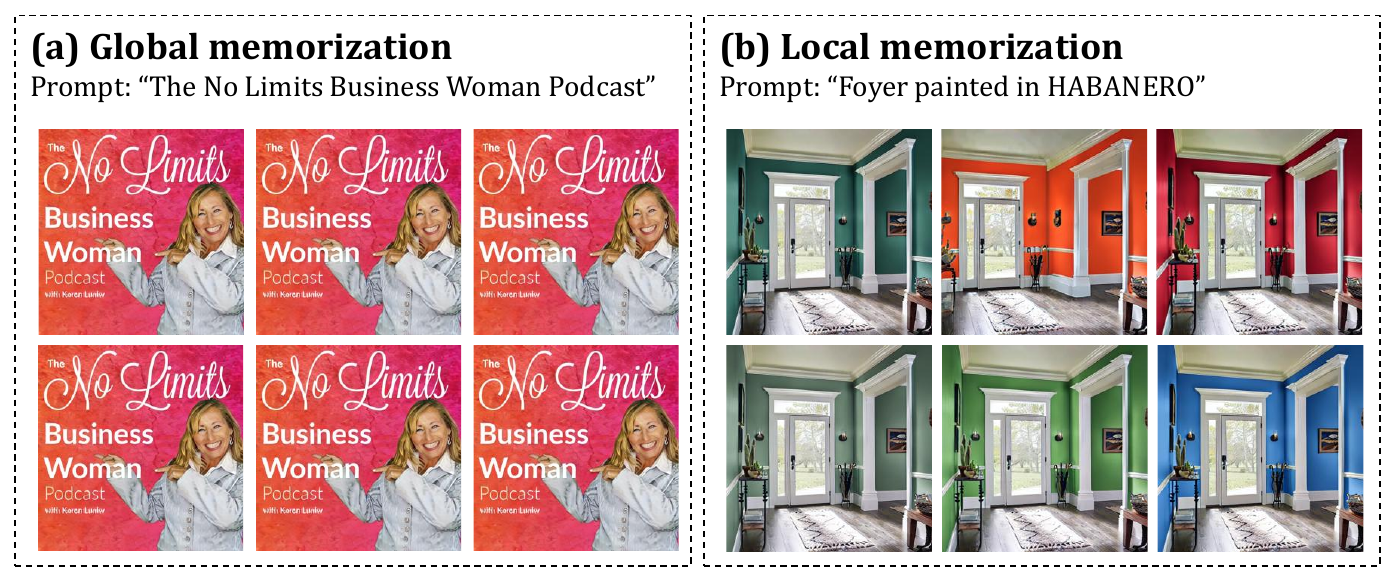}
        \caption{Global and local memorization in Stable Diffusion v1.4 \cite{rombach2022high} across 6 different seeds.}
    \label{fig:different_memo}
\end{figure}

Such memorization can occur globally or locally \cite{chen2025exploring, kowalczuk2025finding}. Global memorization closely reproduces a training image as a whole, while local memorization retains copied content within part of an otherwise varying image. Fig.~\ref{fig:different_memo} illustrates both cases: the global example exhibits whole-image copying, whereas the local example preserves memorized content despite changes in wall color. These variations can make local copying less apparent, highlighting the need to identify its spatial extent. Localization also enables selective mitigation: interventions can target copied regions while preserving surrounding content, reducing the unnecessary changes that image-wide suppression may introduce.

Existing detection methods provide useful evidence of copying, but only limited efforts have been devoted to localizing where memorized content occurs within generated images. Retrieval-based methods compare generated images against training examples~\cite{carlini2023extracting,somepalli2023diffusion,pizzi2022self}; retrieval-free methods analyze model responses, such as conditional-unconditional score differences~\cite{wen2024detecting,ross2025geometric,jeon2024understanding}. These approaches can indicate whether an output is memorized, but a single image-level score does not show which region is responsible. Attention-based methods use cross-attention to identify potentially copied regions~\cite{ren2024unveiling,chen2025exploring}, and coordinate-wise curvature differences offer a complementary geometric signal for localization~\cite{kim2026localizing, joo2026rethinking}. However, a map estimated at one denoising state may include transient responses and background noise. A reliable local detector should therefore capture evidence that persists along the denoising trajectory.

This requirement is closely tied to inference-time mitigation. Prior methods modify prompts or embeddings~\cite{somepalli2023understanding,wen2024detecting}, adjust initial noise~\cite{han2026adjusting}, or change classifier-free guidance (CFG) during sampling~\cite{ho2022classifier,jain2025classifier}. Dynamic attention intervention offers another way to selectively alter text conditioning~\cite{zhao2026you}. Although these methods intervene at different stages of generation, applying their changes globally can alter non-memorized regions alongside copied content. Therefore, a spatial memorization signal can help direct existing interventions toward locally copied regions, suppressing memorization while preserving the surrounding content.

Motivated by these, we present Trajectory-Aware Curvature Observation (\textbf{TACO}), a unified framework for local memorization detection and mitigation. Our empirical study shows that memorization-related curvature becomes pronounced in late denoising and remains spatially consistent across neighboring states. TACO uses this property to estimate prompt-dependent coordinate-wise curvature at multiple late-stage states, refine each response using the latent's spatial structure, and aggregate the observations into a continuous memorization map. The map serves two roles: thresholding it yields a localization mask, while retaining its continuous values provides spatial weights for mitigation. 
In mitigation, TACO separates each intervention into a component aligned with a reference direction, which preserves the baseline objective, and a component that can be spatially adjusted. When the baseline objective gradient is available, TACO uses it as the reference direction, preserving the aligned component and spatially redistributing the orthogonal component according to the memorization map. When it is unavailable, TACO uses the score difference as the reference, combining map-weighted CFG attenuation with a transverse correction orthogonal to that direction.

We evaluate TACO on Stable Diffusion (SD) models \cite{rombach2022high} using benchmarks for local-region detection and memorization mitigation. TACO improves localization over competitive baselines, and its integration with diverse existing mitigation methods advances the Pareto frontier \cite{miettinen1999nonlinear}, achieving a more favorable trade-off between memorization suppression and image quality, as measured by text alignment and perceptual quality.
Our contributions are threefold:
\begin{itemize}
    \item We introduce \textbf{TACO}, a unified framework for memorization localization and mitigation in T2I diffusion models that advances the Pareto frontiers between memorization suppression and generation quality.
    \item We leverage the temporal persistence of late-stage curvature, combining coordinate-wise curvature estimation, latent-guided refinement, and trajectory aggregation to produce spatially coherent memorization maps.
    \item We develop map-guided steering for existing inference-time interventions and validate TACO's localization and mitigation benefits across multiple baselines and Stable Diffusion model variants.
\end{itemize}

\section{Method}
\label{sec:method}

\subsection{Preliminaries}

We consider a latent-space text-to-image diffusion model that generates an image through an iterative denoising process with $T$ sampling steps. Throughout this paper, \emph{step} refers exclusively to an actual denoising update during inference, whereas \emph{timestep} refers to the diffusion-time coordinate used by the scheduler. 

Let
$x_0 \in \mathbb{R}^{C\times H\times W}$
denote the initial Gaussian noise, and let
$x_t  \in \mathbb{R}^{C\times H\times W}$ denote the latent state obtained after $t$ steps.
The resulting sampling trajectory is therefore
$\{x_t\}_{t=0}^{T}$,
where $x_T$ is the final denoised latent that is subsequently decoded into the generated image.

Given a text prompt $p$, at each sampling step $t$, the denoising network produces an unconditional noise prediction
$\epsilon_t^{\mathrm{u}}(x_t)$
and a conditional prediction
$\epsilon_t^{\mathrm{c}}(x_t,p)$.
Classifier-free guidance combines them as
\begin{equation}
\epsilon_t^{\mathrm{g}}(x_t,p)
=
\epsilon_t^{\mathrm{u}}(x_t)
+
s
\left(
\epsilon_t^{\mathrm{c}}(x_t,p)
-
\epsilon_t^{\mathrm{u}}(x_t)
\right),
\end{equation}
where $s$ is the guidance scale.
The scheduler then applies the corresponding denoising transition to obtain the next latent state. Thus, increasing $t$ corresponds to progressing along the inference-time denoising trajectory from $x_0$ toward the final generated latent $x_T$.

Our study considers two closely related tasks: \emph{memorization localization} and \emph{memorization mitigation}.
The localization task identifies regions of the generated image that contain memorized content using signals from the denoising trajectory. The mitigation task uses these regions to guide inference-time interventions, reducing the reproduction of memorized training content while preserving image quality and semantic alignment. For both tasks, TACO produces a continuous spatial memorization map
$M\in\mathbb{R}^{H\times W}$,
where a larger value at spatial location $x$ indicates stronger evidence of memorization-related activity.
For localization, $M$ is converted into a binary mask to determine the spatial extent of memorized content. For mitigation, the same continuous map is retained as a spatial control signal to guide the intervention along the denoising trajectory.

\begin{figure*}[ht]
     \centering
     \includegraphics[width=0.9\textwidth]{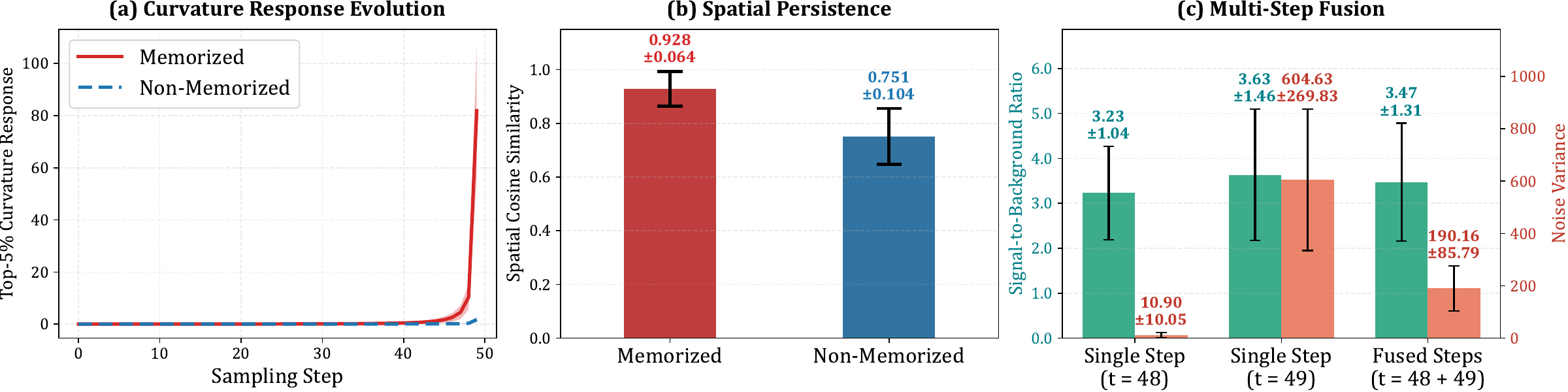}

     \caption{Empirical studies motivating trajectory-aware curvature observation. 
     (a) Evolution of peak curvature intensity along the sampling trajectory. 
     (b) Spatial consistency of curvature responses between neighboring late-stage states. 
     (c) Signal-to-background ratio and background variance of individual and fused late-stage curvature responses.}
     \label{fig:motivation}
\end{figure*}

\subsection{TACO for Memorization Localization}

The key idea of TACO is to identify memorization from curvature patterns that persist along the sampling trajectory. Here, curvature measures the coordinate-wise second-order sensitivity of the denoising residual to the latent state. Given a prompt $p$, TACO observes curvature at multiple states $\{x_t\}_{t\in\mathcal{T}_{\mathrm{obs}}}$ along the trajectory. By preserving spatial correspondence and aggregating responses across these states, TACO captures persistent curvature evidence while reducing the influence of transient responses.

\noindent\textbf{Empirical Evidences for TACO.} We first investigate how curvature responses evolve along the sampling trajectory and whether their spatial patterns remain stable across neighboring states.
All analyses use a sampling process with $T=49$, and are conducted over 200 samples.
For each step, we summarize the curvature map by averaging its strongest $5\%$ spatial responses, which gives a trajectory profile of peak curvature intensity.
As shown in Fig.~\ref{fig:motivation}(a), the curvature response of memorized samples increases sharply toward the end of sampling. In comparison, the response of non-memorized samples increases only mildly and remains substantially lower.
This observation indicates that the curvature associated with memorization becomes particularly pronounced during the late stage of sampling, making these states especially informative for localization.

We next examine whether the spatial locations of strong curvature responses remain stable across neighboring late-stage states.
Specifically, we compute the spatial cosine similarity between the curvature maps at $t=48$ and $t=49$. As shown in Fig.~\ref{fig:motivation}(b), the strong responses associated with memorization tend to persist at consistent spatial locations across neighboring states, providing a reliable temporal correspondence for trajectory-level aggregation.

Finally, we investigate whether combining these neighboring observations improves the stability of the resulting spatial signal.
We average the curvature maps at $t=48$ and $t=49$ and measure their signal-to-background ratio (SBR), where the strongest $10\%$ of responses define the signal region and the remaining locations form the background.
We additionally measure the background variance to quantify fluctuations outside the signal region. In Fig.~\ref{fig:motivation}(c), the fused response maintains a high SBR while substantially reducing the background variance compared with the highly responsive yet noisy observation at $t=49$, providing a favorable balance between signal contrast and background stability.

Together, these findings show that memorization-related curvature strengthens and becomes spatially consistent in late-stage sampling, while aggregation across neighboring states improves background stability. This motivates TACO to model curvature as a trajectory-level spatial signal.

\noindent\textbf{Trajectory-Aware Curvature Signal.} Building on coordinate-wise curvature differences \cite{kim2026localizing}, we first extract the prompt-dependent curvature response at each observation state.
For a given state $x_t$, we isolate the prompt-dependent component of the denoising dynamics as
\begin{equation}
r_t(x_t,p)
=
\epsilon_t^{\mathrm{c}}(x_t,p)
-
\epsilon_t^{\mathrm{u}}(x_t).
\end{equation}

We then estimate the diagonal curvature of the normalized residual $r_t/\sigma_t$, where $\sigma_t=\sqrt{1-\bar{\alpha}_t}$ is the noise scale at state $t$. We use $J$ structured Walsh-Hadamard probes $\{v^{(j)}\}_{j=1}^{J}$ ~\cite{bekas2007estimator}, with fixed spatial assignments,
channel-dependent offsets, and a shared random sign field. These probes provide approximately orthogonal coverage of the latent coordinates. The resulting curvature estimator is
\begin{equation}
\widehat{H}_t
=
\frac{1}{J}
\sum_{j=1}^{J}
v^{(j)}
\odot
\nabla_{x_t}
\left\langle
-\frac{r_t(x_t,p)}{\sigma_t},
v^{(j)}
\right\rangle,
\end{equation}
where $\widehat{H}_t(x,c)$ denotes the curvature at spatial location $x$ and channel $c$. We retain the negative response
$
H_t^{-}(x,c)
=
\operatorname{ReLU}\!\left(-\widehat{H}_t(x,c)\right)
$, and average it across channels to obtain the spatial response
$
C_t(x)
=
\frac{1}{C}
\sum_{c=1}^{C}H_t^{-}(x,c)
$.

To emphasize salient curvature responses, we standardize $C_t$ across spatial locations and apply sigmoid-based reweighting as ${C}_t(x)
=
C_t(x)\operatorname{sigmoid}
\left(
\beta\frac{C_t(x)-\mu_t}{\sigma_t+\epsilon}
\right)$, 
where $\mu_t$ and $\sigma_t$ denote the spatial mean and standard deviation of $C_t$, respectively, and $\beta = 2.0$ controls the strength of the reweighting. This operation enhances relatively strong responses while retaining their original magnitudes.

The key difference from single-state curvature localization is that TACO preserves the spatial correspondence of $C_t$ across the sampling trajectory.
Specifically, for the observation states $\mathcal{T}_{\mathrm{obs}}$, we obtain the sequence
$\{C_t\}_{t\in\mathcal{T}_{\mathrm{obs}}}$ and retain each response at its original spatial location.
This allows curvature evidence to be evaluated jointly across neighboring states, following the trajectory-level persistence observed in our empirical analysis.

\noindent\textbf{Latent-Guided Spatial Refinement.} Although observations across neighboring states provide temporal evidence, each state-wise curvature map may still contain isolated spatial peaks. We therefore refine each response $C_t$ using the spatial structure of its corresponding latent state $x_t$ before trajectory-level aggregation.

Let $\mathbf{f}_t(x)\in\mathbb{R}^{C}$ denote the latent feature at spatial location $x$. We perform Principal Component Analysis (PCA) over all spatial locations and project the centered features onto the first principal direction $\mathbf{u}_t$ as $G_t(x)
=
\mathbf{u}_t^\top
\left(
\mathbf{f}_t(x)-\boldsymbol{\mu}_t
\right)$, where $\boldsymbol{\mu}_t$ is the spatial mean of the latent features. The resulting map $G_t$ summarizes the dominant spatial variation in $x_t$ and serves as a lightweight structural cue. For each $y\in\mathcal{N}(x)$, we define the edge-aware affinity $w_t(x,y)
=
\exp\left(
-\frac{\|x-y\|_2^2}{2\sigma_s^2}
\right)
\exp\left(
-\frac{|G_t(x)-G_t(y)|^2}{2\sigma_g^2(x)}
\right)$, where $\sigma_s$ controls the spatial extent of the neighborhood and $\sigma_g(x)$ controls sensitivity to local variations in the guidance map. We then refine the curvature response by normalized local aggregation:
\begin{equation}
\widetilde{C}_t(x)
=
\frac{
\sum_{y\in\mathcal{N}(x)}
w_t(x,y)C_t(y)
}{
\sum_{y\in\mathcal{N}(x)}
w_t(x,y)
}.
\end{equation}

This refinement suppresses isolated peaks that lack local support, preserves spatially consistent responses, and limits smoothing across latent boundaries. The refined map $\widetilde{C}_t$ is subsequently used for trajectory-level aggregation.

\noindent\textbf{Trajectory-Level Memorization Map.} Fig.~\ref{fig:motivation}(a) shows that memorization-related curvature becomes stronger during the late denoising stages. Based on this observation, we select a short sequence of neighboring late-stage states for curvature observation, which concentrates computation on the stages with the strongest evidence while avoiding initial states that provide limited additional information.

Let $\mathcal{T}_{\mathrm{obs}}$ denote the selected observation states, with
$t_{\mathrm{start}}$ and $t_{\mathrm{end}}$ denoting the earliest and latest
states, respectively. We aggregate their refined curvature responses using
a normalized weighted sum:
\begin{equation}
M(x)
=
\frac{
\sum_{t\in\mathcal{T}_{\mathrm{obs}}}
w_t\widetilde{C}_t(x)
}{
\sum_{t\in\mathcal{T}_{\mathrm{obs}}}w_t
},
\,
w_t
=
2-\frac{t-t_{\mathrm{start}}}
{t_{\mathrm{end}}-t_{\mathrm{start}}}.
\end{equation}

This weighting scheme assigns greater importance to earlier observations within the selected interval. The motivation is that neighboring states exhibit complementary characteristics as in Fig.~\ref{fig:motivation}(c): earlier observations can provide more stable spatial responses with lower background noises, whereas later observations can exhibit stronger curvature contrast alongside increased background variation.

Trajectory aggregation yields a continuous spatial map $M$ that summarizes temporally persistent and spatially coherent curvature evidence across the late-stage states. TACO uses this map in two ways: thresholding $M$ provides a binary localization mask, while its continuous values serve as spatial weights for inference-time mitigation.

\subsection{TACO-Steered Memorization Mitigation}

The continuous TACO map $M$ provides a spatial signal for steering an existing inference-time memorization mitigation method. At each applicable state, TACO takes the intervention generated by the baseline method and spatially modulates its effect toward regions with stronger memorization responses, while preserving the baseline method's original mitigation objective. 

Let $\Delta_t^{\mathrm{base}}$ denote the intervention produced by the baseline method at sampling state $t$, and let $g_t$ denote the gradient associated with its mitigation objective.
We decompose the baseline intervention into an objective-aligned component and an orthogonal tangent component:
\begin{equation}
\Delta_t^{\mathrm{base}}
=
\Delta_t^{\parallel}
+
\Delta_t^{\perp},
\end{equation}
where $\Delta_t^{\parallel}
=
\frac{
\left\langle
\Delta_t^{\mathrm{base}},g_t
\right\rangle
}{
\|g_t\|_2^2+\epsilon
}
g_t
$, and $
\Delta_t^{\perp}
=
\Delta_t^{\mathrm{base}}
-
\Delta_t^{\parallel}$. $\Delta_t^{\parallel}$ captures the part of the intervention directly aligned with the original mitigation objective, while $\Delta_t^{\perp}$ provides the degrees of freedom that can be spatially redistributed.

TACO applies the memorization map $M$ only to this redistributable component $\Delta_t^{\perp}$.
At step $t$, $M$ is first resized to the spatial resolution of $x_t$ and locally expanded to tolerate small localization errors: $
M_t^{\mathrm{dil}}
=
\operatorname{MaxPool}
\left(
\operatorname{Resize}(M)
\right)$. We then normalize $M_t^{\mathrm{dil}}$ to 
$
\widetilde{M}_t(x)
=
\frac{
M_t^{\mathrm{dil}}(x)-M_t^{\min}
}{
M_t^{\max}-M_t^{\min}+\epsilon
}$. The resulting weight assigns stronger intervention to locations with stronger memorization evidence while maintaining a continuous transition across neighboring regions.

Afterwards, $\Delta_t^{\perp}$ is spatially reweighted according to $\widetilde{M}_t(x)$:
\begin{equation}
\widehat{\Delta}_t^{\perp}
=
\widetilde{M}_t
\odot
\Delta_t^{\perp}.
\end{equation}

Since spatial reweighting can change the magnitude of the intervention, TACO rescales $\widehat{\Delta}_t^{\perp}$ to preserve the magnitude of the original tangent update as: 
\begin{equation}
\Delta_t^{\perp,\mathrm{TACO}}
=
\frac{
\|\Delta_t^{\perp}\|_2
}{
\|\widehat{\Delta}_t^{\perp}\|_2+\epsilon
}
\widehat{\Delta}_t^{\perp}.
\end{equation} 

The final TACO-steered intervention is therefore $
\Delta_t^{\mathrm{TACO}}
=
\Delta_t^{\parallel}
+
\Delta_t^{\perp,\mathrm{TACO}}$.

This formulation preserves the component required by the original mitigation objective and the overall magnitude of the redistributable intervention, while concentrating its spatial effect according to the TACO memorization map.
Consequently, TACO changes where the mitigation acts rather than simply adjusting its global strength.

More generally, the steering operation is independent of the specific form of the underlying mitigation procedure. When the objective gradient $g_t$ is unavailable, TACO uses the current score difference $r_t=\epsilon_t^{\mathrm{c}}-\epsilon_t^{\mathrm{u}}$ as the reference direction for gradient decomposition. Specifically, TACO computes the gradient of a map-weighted attraction objective and projects it onto the subspace orthogonal to $r_t$, yielding a transverse correction that is normalized and scaled before being added to the baseline intervention. In parallel, the TACO map locally attenuates classifier-free guidance in regions with stronger memorization responses. These two components jointly steer the mitigation toward localized memorized content while preserving the orthogonality of the gradient correction to the current score-difference direction, without requiring an explicit gradient from the underlying mitigation objective.

Overall, TACO establishes a direct connection between localization and mitigation: the same continuous trajectory-aware map that identifies where memorization is expressed is reused to determine where the mitigation intervention should be concentrated. Detailed algorithms for localization and mitigation are provided in the supplementary material.

\section{Experiments}

\subsection{Implementation Details}

We evaluate memorization localization on SD v1.4 and v2.1 \cite{rombach2022high}, and memorization mitigation on SD v1.4 and v2.0, following the configurations of the corresponding benchmarks. We use DDIM sampling \cite{song2020denoising} with $50$ inference steps and a classifier-free guidance scale of $7.5$. TACO observes the last two denoising states, $\mathcal{T}_{\mathrm{obs}}=\{48,49\}$, and estimates coordinate-wise curvature using 16 structured Walsh-Hadamard probes. All experiments are and conducted on one NVIDIA GeForce RTX 4090 GPU under Ubuntu 22.04. 

\subsection{Benchmarks and Baselines}

\noindent\textbf{Memorization Localization.}
We evaluate localization on the benchmark provide by~\cite{kim2026localizing} under two settings: \textit{TV only}, containing samples with localized memorization, and \textit{All}, which additionally includes non-memorized samples. Using ground-truth binary masks, we report mean Intersection over Union (mIoU), which measures spatial overlap with memorized regions, and pixel-level accuracy (Acc), which measures the proportion of correctly classified pixels. Higher values indicate better localization. 

Baselines include Bright Ending (\texttt{BE})~\cite{chen2025exploring} and four variants from~\cite{kim2026localizing}: curvature-difference estimators and their score-difference approximations, using either the unconditional model (\texttt{cov}, \texttt{score\_diff}) or a less-trained model as reference (\texttt{cov\_bad}, \texttt{score\_diff\_bad}).

\begin{table}[t]
\caption{Quantitative evaluation of memorization localization. Bold and underlined values denote the best and second-best results, respectively.}
\label{tab:localization}
\centering
\resizebox{\linewidth}{!}{%
\begin{tabular}{l l cc cc}
\toprule
\multirow{2}{*}{Model} & \multirow{2}{*}{Method} & \multicolumn{2}{c}{\textbf{TV only}} & \multicolumn{2}{c}{\textbf{All}} \\
\cmidrule(lr){3-4} \cmidrule(lr){5-6}
& & mIoU($\uparrow$) & ACC($\uparrow$) & mIoU($\uparrow$) & ACC($\uparrow$) \\
\midrule
\multirow{8}{*}{SD v1.4}
& All-zeros  & $0.000$ & $0.437$ & $0.318$ & $0.614$ \\
& All-ones   & $0.563$ & $0.563$ & $0.386$ & $0.386$ \\
\cmidrule{2-6}
& \texttt{BE}         & $0.693$ & $0.766$ & $0.467$ & $0.780$ \\
& \texttt{cov  }      & $0.907$ & $0.941$ & $\underline{0.907}$ & $0.937$ \\
& \texttt{cov\_bad } & \underline{$0.922$} & \underline{$0.950$} &$0.847$ & \underline{$0.940$} \\
& \texttt{score\_diff  }   & $0.826$ & $0.891$ & $0.847$ & $0.906$ \\
& \texttt{score\_diff\_bad} & $0.857$ & $0.911$ & $0.645$ & $0.914$ \\ \cmidrule{2-6}
& \textbf{Ours} & $\mathbf{0.942}$ & $\mathbf{0.964}$ & $\mathbf{0.926}$ & $\mathbf{0.949}$ \\
\midrule
\multirow{8}{*}{SD v2.1}
& All-zeros  & $0.000$ & $0.352$ & $0.239$ & $0.507$ \\
& All-ones   & $0.648$ & $0.648$ & $0.493$ & $0.493$ \\
\cmidrule{2-6}
& \texttt{BE}         & $0.794$ & $0.852$ & $0.815$ & $0.887$ \\

& \texttt{cov} & $0.938$ & $0.958$ & \underline{$0.908$} & $0.968$ \\
& \texttt{cov\_bad} & \underline{$0.942$} & \underline{$0.963$} & $0.902$ & \underline{$0.971$} \\
& \texttt{score\_diff}  & $0.781$ & $0.836$ & $0.814$ & $0.876$ \\
& \texttt{score\_diff\_bad} & $0.915$ & $0.944$ & $0.899$ & $0.957$ \\ \cmidrule{2-6}
&  \textbf{Ours} & $\mathbf{0.957}$ &$\mathbf{0.972}$ & $\mathbf{0.923}$ & $\mathbf{0.977}$ \\
\bottomrule
\end{tabular}%
}
\end{table}

\noindent\textbf{Memorization Mitigation.}
We evaluate memorization mitigation on the benchmark provided by~\cite{webster2023reproducible}, which covers both local and global memorization. We use SSCD \cite{pizzi2022self} to measure the visual similarity between each mitigated image and its corresponding reference, where lower similarity indicates stronger memorization suppression. To assess whether this suppression preserves generation quality, we additionally use CLIP \cite{radford2021learning} to measure image-text alignment and ImageReward \cite{xu2023imagereward} to predict prompt-conditioned human preference. Higher CLIP and ImageReward scores indicate better semantic alignment and perceptual quality, respectively. We integrate TACO into four baselines: batch-wise and sample-wise initial-noise adjustment \cite{han2026adjusting} (\texttt{Init Batch}, \texttt{Init Sample}), and early-stage CFG perturbations with zero or opposite guidance \cite{jain2025classifier} (\texttt{Basin Zero}, \texttt{Basin Neg}). 
We vary each baseline's mitigation strength and compare its Pareto frontiers with and without TACO in the SSCD-CLIP and SSCD-ImageReward spaces to assess the memorization-quality trade-off. Unless otherwise specified, we follow the default configurations of each baseline.

\begin{figure*}[ht]
     \centering
     \includegraphics[width=0.9\textwidth]{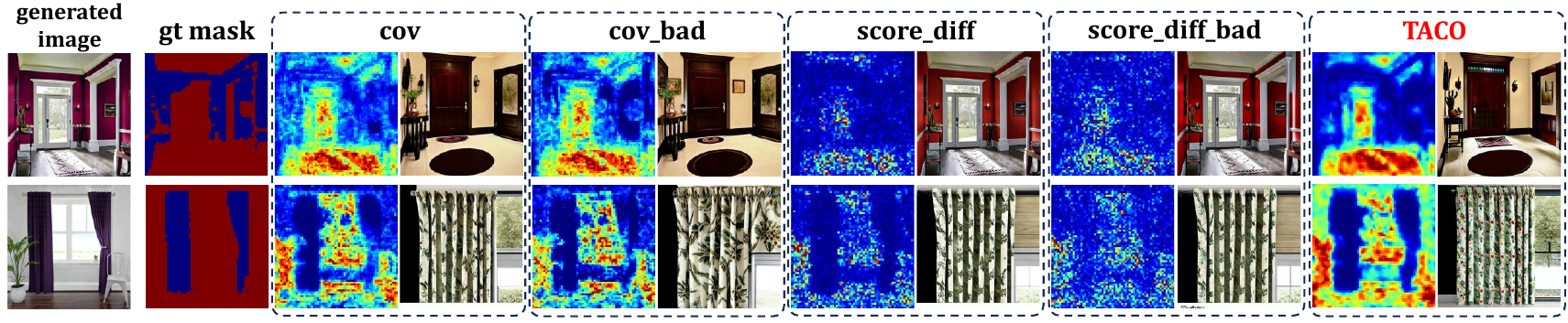}
     \caption{     Qualitative comparison of memorization localization and map-guided mitigation for prompts: ``Foyer painted in HABANERO'' (top) and ``Plymouth Curtain Panel featuring Madelyn - White Botanical Floral Large Scale by heatherdutton'' (bottom). For each method, the left panel shows its original localization map, with blue and red indicating weaker and stronger memorization responses, respectively. The right panel shows the mitigated image obtained by applying our proposed steering method with the corresponding localization map.
     }
     \label{fig:detection_vis}
\end{figure*}

\subsection{Experimental Results}

\noindent\textbf{Memorization Localization.} Tab.~\ref{tab:localization} reports quantitative localization results on SD v1.4 and SD v2.1 under the TV only and All settings. TACO consistently achieves the best overall performance across both models and evaluation settings, demonstrating its ability to localize memorized regions accurately while distinguishing them from non-memorized content. On SD v1.4, TACO achieves an mIoU of $0.942$ and an accuracy of $0.964$ on TV only, improving over the strongest competing results by $2.0$ and $1.4$ percentage points, respectively. On All, it further outperforms the best competing results by $1.9$ and $0.9$ percentage points. The improvements remain consistent on SD v2.1. 

TACO also maintains a competitive computational cost. Averaged over individual samples, \texttt{cov}, \texttt{cov\_bad}, \texttt{score\_diff}, and \texttt{score\_diff\_bad} require $1.87$, $2.40$, $1.32$, and $2.57$ seconds, respectively, while TACO requires $2.41$ seconds. Although TACO aggregates curvature evidence from two late-stage states, its runtime is nearly identical to that of \texttt{cov\_bad}, while achieving markedly better localization performance. 


Fig.~\ref{fig:detection_vis} provides a qualitative comparison between TACO and the four baselines evaluated in~\cite{kim2026localizing}. TACO produces more spatially coherent responses with sharper boundaries, better capturing memorized regions while suppressing diffuse responses in surrounding non-memorized areas. To further examine the utility of these maps for mitigation, we apply our proposed steering pipeline to \texttt{Init Batch} using each method's predicted map. 
The resulting images demonstrate that the more precise localization provided by TACO enables more targeted intervention while better preserving image quality. In contrast, less precise baseline maps may lead to suboptimal mitigation results. \textbf{This comparison indicates that improved localization not only enhances interpretability but also provides more effective spatial guidance for memorization mitigation.}

\begin{figure*}[ht]
    \centering
    \includegraphics[width=0.9\textwidth]{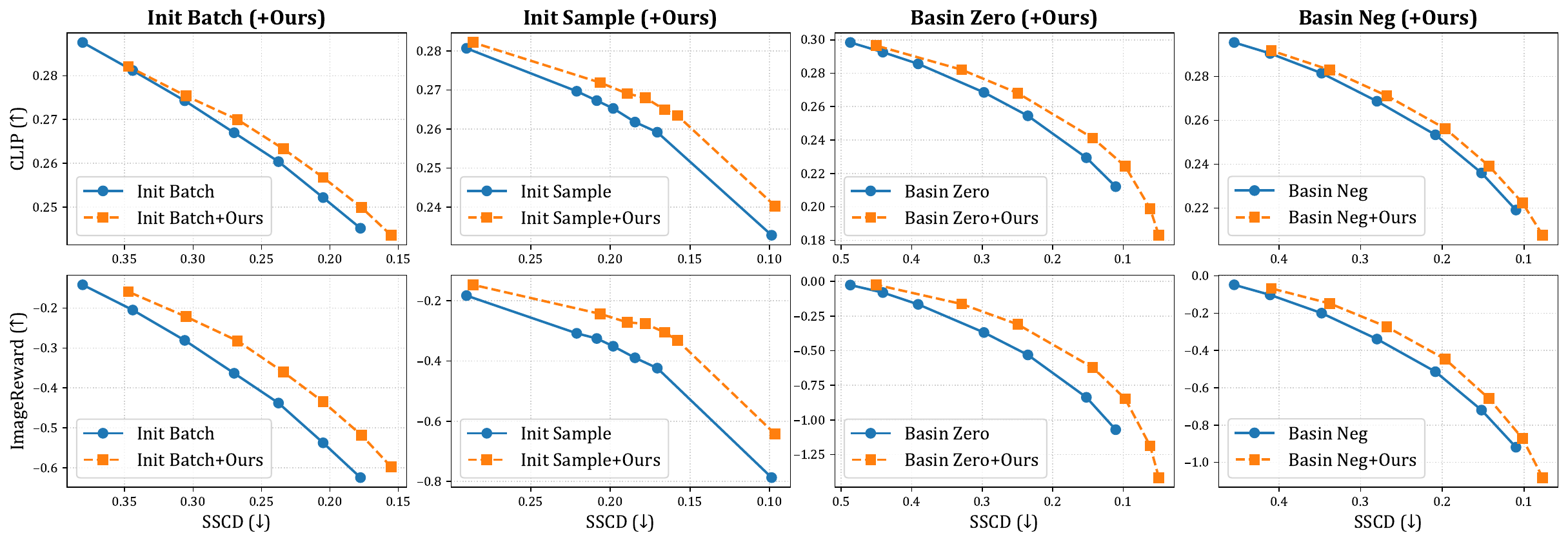}
    \caption{SSCD-CLIP and SSCD-ImageReward Pareto fronts in SD v1.4.}
    \label{fig:pareto_v1}
\end{figure*}

\begin{figure*}[ht]
    \centering
    \includegraphics[width=0.9\textwidth]{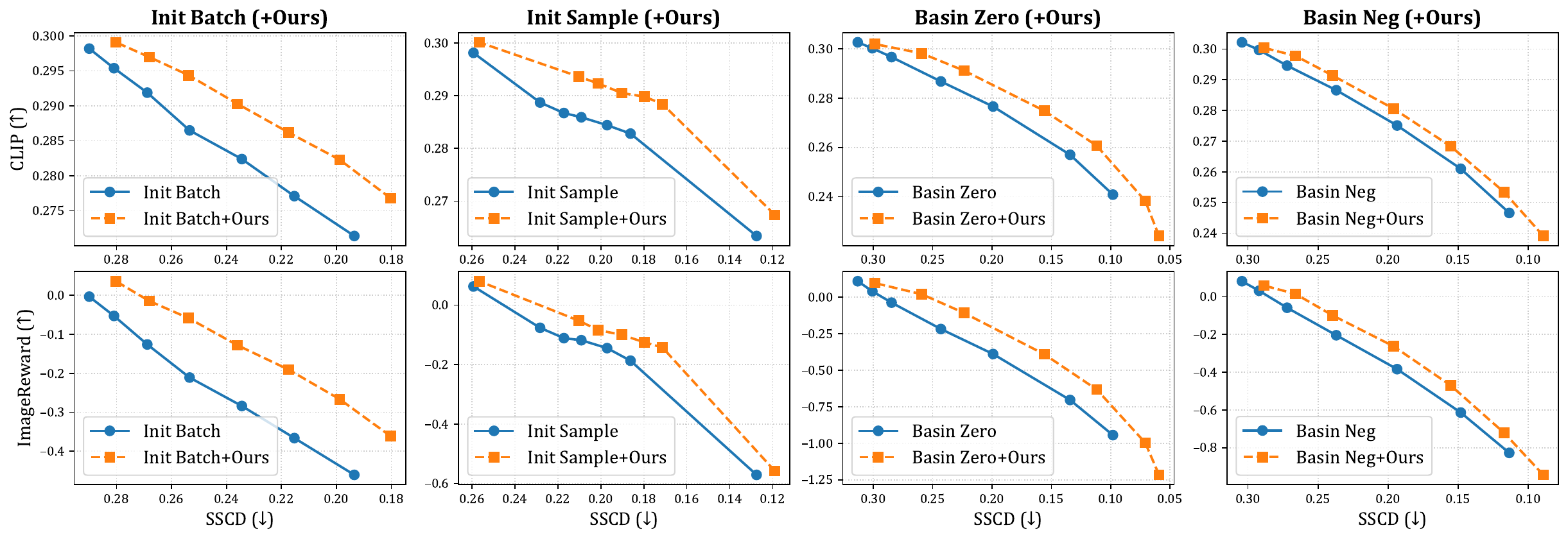}
    \caption{SSCD-CLIP and SSCD-ImageReward Pareto fronts in SD v2.0.}
    \label{fig:pareto_v2}
\end{figure*}

\noindent\textbf{Memorization Mitigation.} We next evaluate whether TACO can improve existing inference-time mitigation methods without sacrificing semantic alignment or perceptual quality. Fig.~\ref{fig:pareto_v1} and Fig.~\ref{fig:pareto_v2} compare the Pareto frontiers of the original baselines and their TACO-steered counterparts on SD v1.4 and SD v2.0, respectively. Each point corresponds to a different mitigation strength. Across both diffusion models, TACO steering consistently improves the trade-off between memorization suppression and generation quality. In the SSCD-CLIP space, the TACO-steered frontiers extend toward lower SSCD and higher semantic alignment; similar improvements are observed in the SSCD-ImageReward space. Importantly, these gains are observed across four distinct intervention strategies, including batch-wise and sample-wise initial-noise adjustment and early-stage CFG perturbations with zero or opposite guidance. This consistency indicates that TACO is not tied to a particular mitigation objective, but can spatially redirect existing interventions toward memorization-related regions. The quantitative Pareto-front statistics in the supplementary material further support this observation.

\begin{figure*}[t]
    \centering
    \includegraphics[width=0.75\textwidth]{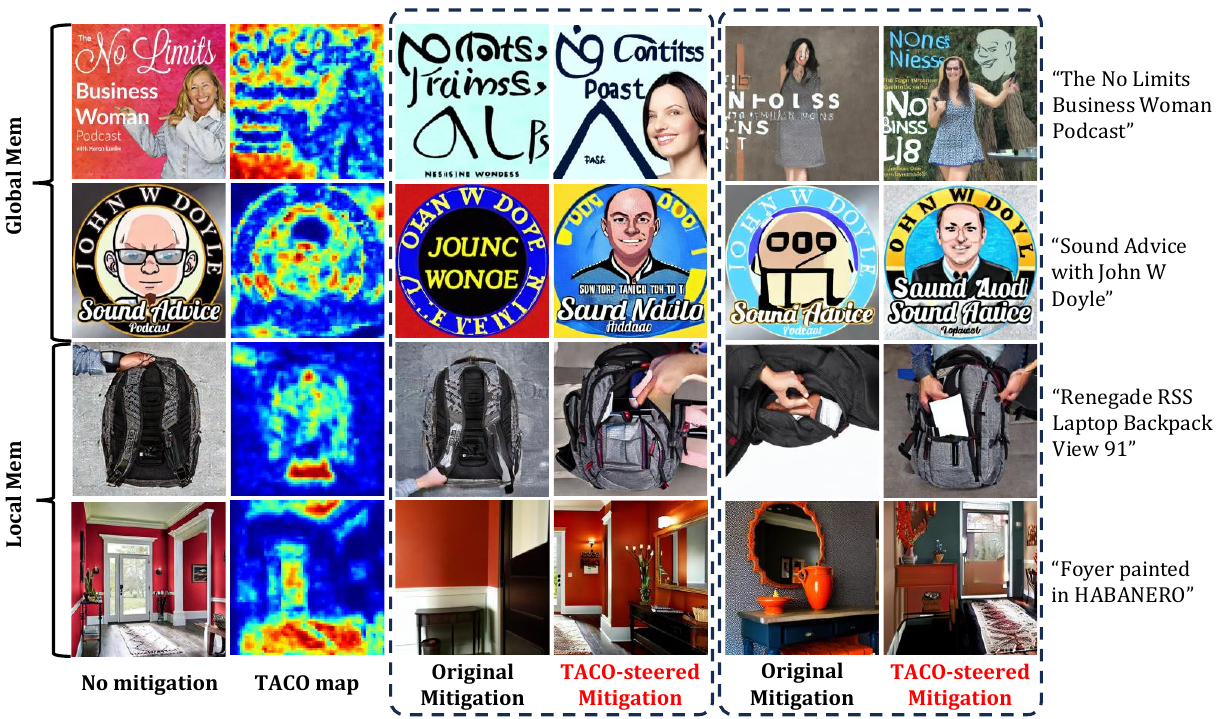}
    \caption{Qualitative comparison of existing memorization mitigation methods with and without TACO steering. Each pair compares the original mitigation baseline with its TACO-steered counterpart using the same generation seed, ensuring a controlled comparison. }
    \label{fig:mitigation_vis}

\end{figure*}

Fig.~\ref{fig:mitigation_vis} further illustrates the qualitative benefits of TACO steering. Although TACO derives a spatial map for memorization localization, \textbf{its steering mechanism is not limited to local memorization and can also improve the mitigation of global memorization.} Compared with baseline methods, TACO better suppresses memorized content while preserving prompt-specific semantic details. In the second example, it preserves the identity-related depiction of ``John Doyle'', whereas the baseline simplifies the subject into text. In the fourth example, TACO better retains the structural details of the ``foyer'', which are less clearly preserved by the baseline. These results highlight the broader applicability of TACO steering: it guides existing mitigation methods to suppress memorized content while better retaining relevant visual and semantic information, whether memorization affects a localized region or the image as a whole.

Overall, the quantitative and qualitative results show that TACO provides a unified mechanism for both memorization localization and mitigation: it produces more precise spatial maps and uses them to improve the memorization-quality trade-off while retaining important prompt-specific details.

\subsection{Ablation Study and Sensitivity Analysis}

\noindent \textbf{Ablation Study of TACO components.} We conduct ablation experiments to evaluate the contributions of trajectory-aware aggregation and local refinement. Tab.~\ref{tab:taco_ablation} reports the results on SD v1.4 and SD v2.1 under both TV only and All settings. Removing trajectory aggregation and using a single observation step consistently reduces localization performance on TV only, indicating that combining curvature evidence across adjacent late steps provides more reliable spatial cues. Local refinement provides complementary improvements by sharpening the spatial localization of memorized regions. Removing this module decreases the performance, with a particularly noticeable mIoU drop on the All split of SD v2.1, from $0.923$ to $0.905$. This suggests that trajectory aggregation improves the reliability of memorization evidence, while local refinement helps translate that evidence into more accurate spatial boundaries.

\begin{table}[t]
\caption{Ablation study of TACO components.}
\small
\label{tab:taco_ablation}
\centering
\resizebox{0.9\linewidth}{!}{%
\begin{tabular}{l l cc cc}
\toprule
\multirow{2}{*}{Model} & \multirow{2}{*}{Method} & \multicolumn{2}{c}{\textbf{TV only}} & \multicolumn{2}{c}{\textbf{All}} \\
\cmidrule(lr){3-4} \cmidrule(lr){5-6}
& & mIoU($\uparrow$) & ACC($\uparrow$) & mIoU($\uparrow$) & ACC($\uparrow$) \\
\midrule
\multirow{4}{*}{SD v1.4}
& \textbf{Ours} & $\mathbf{0.942}$ & $\mathbf{0.964}$ & $\mathbf{0.926}$ & $\mathbf{0.949}$ \\ \cmidrule{2-6}
& w/o Traj (only $48$) & $0.933$ & $0.959$ & $0.918$ & $0.945$ \\
& w/o Traj (only $49$) & $0.936$ & $0.960$ & $0.922$ & $0.947$ \\
& w/o Local Refine & \underline{$0.939$} & \underline{$0.962$} & \underline{$0.923$} & \underline{$0.948$} \\
\midrule
\multirow{4}{*}{SD v2.1}
& \textbf{Ours} & $\mathbf{0.957}$ & $\mathbf{0.972}$ & \underline{$0.923$} & $\mathbf{0.977}$ \\ \cmidrule{2-6}
& w/o Traj (only $48$) & $0.949$ & $0.966$ & $0.894$ & $0.970$ \\
& w/o Traj (only $49$) & \underline{$0.951$} & \underline{$0.967$} & $\mathbf{0.926}$ & \underline{$0.973$} \\
& w/o Local Refine & $0.949$ & $0.966$ & $0.905$ & $0.972$ \\
\bottomrule
\end{tabular}%
}
\end{table}

\begin{figure}[t]
    \centering
    \includegraphics[width=0.9\linewidth]{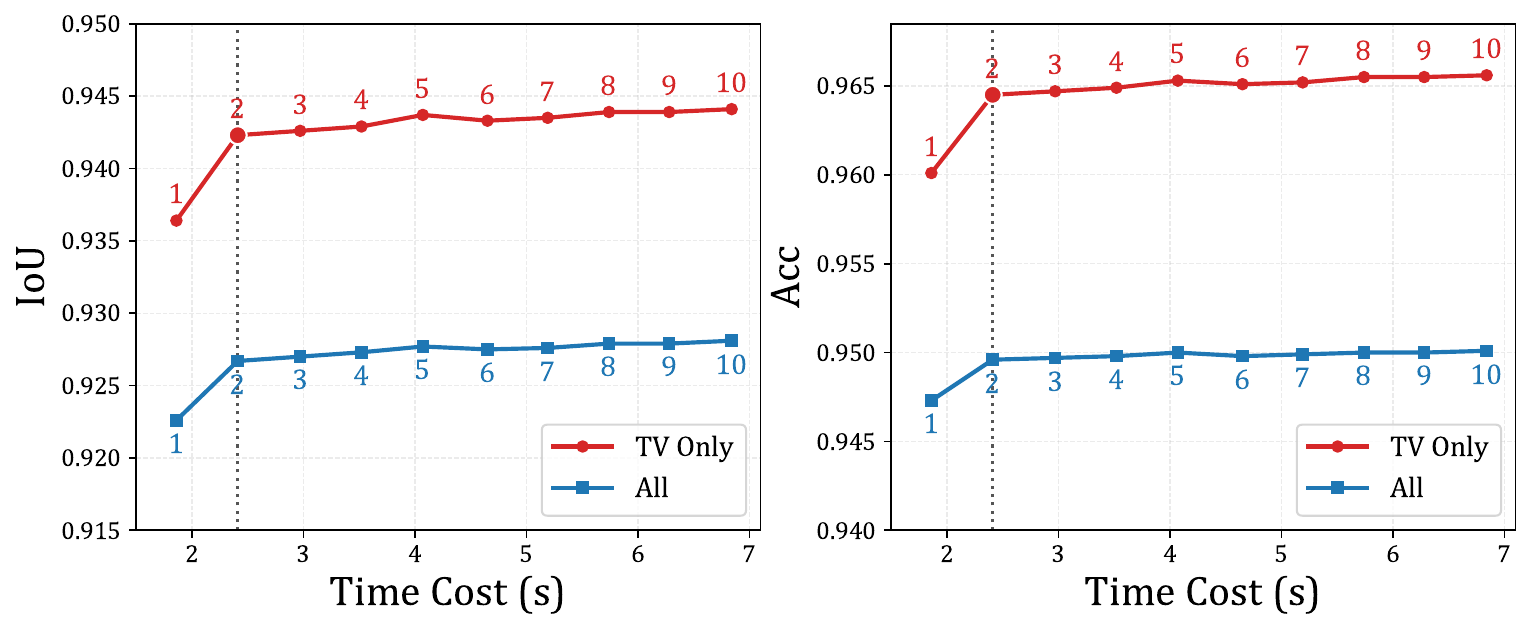}
    \caption{Sensitivity analysis of the observation-window size $|\mathcal{T}_{\mathrm{obs}}|$. The numbers annotated around each point indicate the corresponding window size.}
    \label{fig:sensitivity_t}
\end{figure}

\noindent\textbf{Sensitivity to the Observation-Window Size $|\mathcal{T}_{\mathrm{obs}}|$.}
We study how the number of observation states affects TACO. Starting from the latest denoising states, we progressively expand the observation window toward earlier states. With sampling steps indexed from $0$ to $49$, the additional states are introduced in reverse order, such as $47$, then $46$, and so on. Thus, increasing $|\mathcal{T}_{\mathrm{obs}}|$ adds earlier states for curvature computation while retaining the previously selected late-stage observations.
As shown in Fig.~\ref{fig:sensitivity_t}, increasing the number of observed steps leads to an approximately linear increase in computation time, since curvature estimation must be performed at more points along the denoising trajectory. In contrast, the gains in mIoU and pixel-level accuracy diminish as the window expands. This trend is consistent with the observation that informative curvature changes are concentrated in the late denoising steps: adding earlier steps with less information incurs additional cost but contributes progressively less useful localization evidence. We therefore use $|\mathcal{T}_{\mathrm{obs}}|=2$, corresponding to steps $\{48, 49\}$, as a practical balance between localization accuracy and computational efficiency.

\begin{figure}[t]
    \centering
    \includegraphics[width=0.9\linewidth]{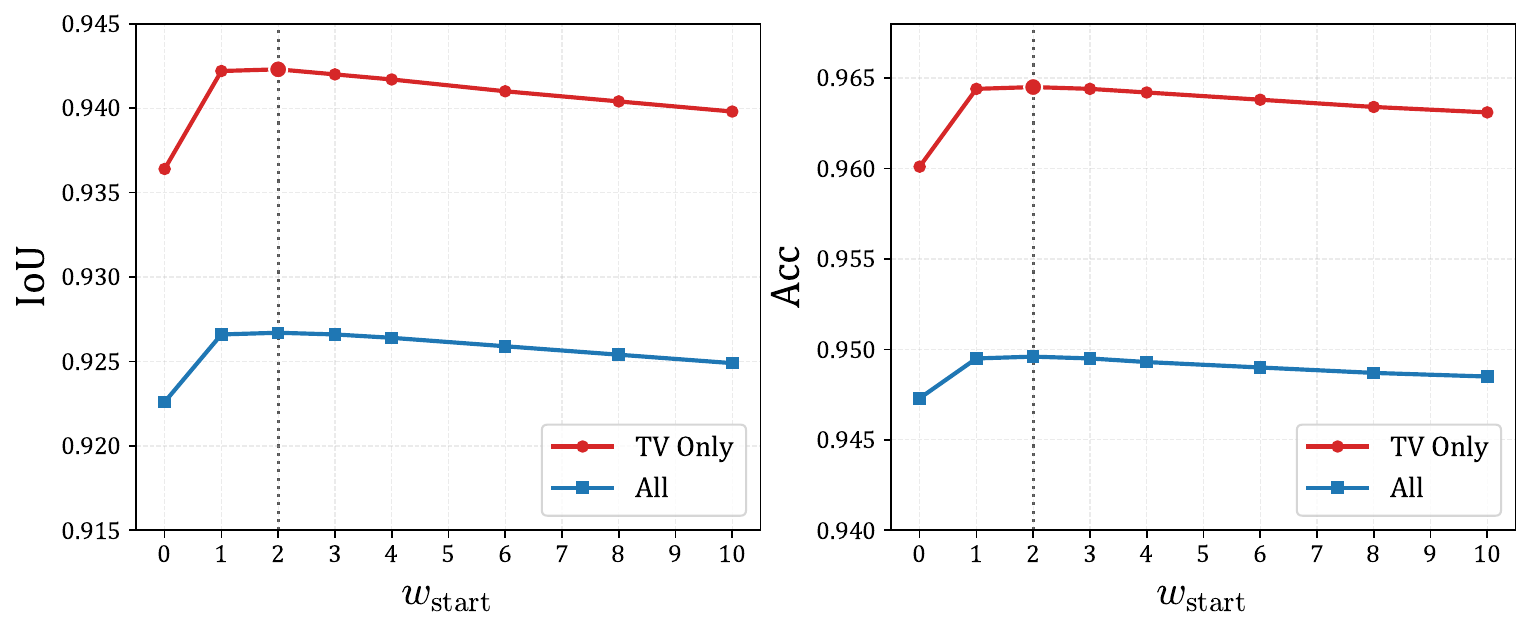}
    \caption{Sensitivity analysis of the aggregation weight $w_\mathrm{start}$.}
    \label{fig:sensitivity_w}

\end{figure}

\noindent\textbf{Sensitivity to the Aggregation Weight.}
For $\mathcal{T}_{\mathrm{obs}}=\{48,49\}$, we vary the weight $w_{t_{\mathrm{start}}}$ assigned to $t=48$, while fixing $w_{t_{\mathrm{end}}}=1$ for the later state $t=49$. Setting $w_{t_{\mathrm{start}}}=0$ retains only the response at the later state. As shown in Fig.~\ref{fig:sensitivity_w}, increasing $w_{t_{\mathrm{start}}}$ initially improves localization by incorporating the more spatially stable evidence from the earlier state. However, excessive weighting reduces the contribution of the stronger curvature contrast at the later state, leading to a decline in performance. The best results are obtained with $w_{t_{\mathrm{start}}}=2$, highlighting the importance of balancing early-state spatial stability with late-state curvature sensitivity.

\section{Conclusion}

In this paper, we present TACO, a trajectory-aware curvature observation framework for local memorization detection and mitigation in diffusion models. TACO aggregates curvature responses across neighboring late denoising steps and refines the resulting spatial evidence using latent-guided local refinement. The resulting continuous memorization map provides a unified interface for both tasks: it identifies memorized regions for localization and spatially guides existing inference-time interventions for mitigation. Extensive experiments demonstrate the effectiveness of TACO. Qualitative results further show that TACO helps preserve prompt-specific semantic details while directing mitigation toward memorization-related regions. Our findings highlight the potential of trajectory-aware spatial analysis as a unified approach to understanding and mitigating memorization in diffusion models.

\noindent\textbf{Limitations and Future Works.}
Despite its effectiveness, estimating coordinate-wise curvature through vector-Jacobian products at multiple states introduces additional inference-time computation. Although TACO observes only a small number of late-stage states, this overhead may still limit its use in latency-sensitive applications. Future works could explore more efficient curvature estimators or probe-sharing strategies to reduce computational cost while retaining accurate localization and effective spatial steering.

{
    \small
    \bibliographystyle{ieeenat_fullname}
    \bibliography{main}
}

\clearpage
\setcounter{page}{1}
\maketitlesupplementary

\begin{appendix}

\section{Overview}

This supplementary material provides background related works, methodological details, and more experimental details and results. Specifically:
\begin{itemize}
    \item Sec.~\ref{SecA} reviews prior works on diffusion-model memorization, its localization, and its mitigation.
    \item Sec.~\ref{SecB} presents the pseudocode for TACO localization and steering, together with an analysis of its computational overhead.
    \item Sec.~\ref{SecC} describes the experimental setup and reports additional quantitative and qualitative results, including quantitative Pareto-front comparisons, and more examples of memorization localization and TACO-steered mitigation.
\end{itemize}

\section{Related Works}
\label{SecA}

\subsection{Memorization in Diffusion Models}

Diffusion models generate images by reversing a progressive noising process~\cite{ho2020denoising,song2021scorebased,songdenoising}. In text-to-image (T2I) systems, latent diffusion combines this process with text conditioning~\cite{rombach2022high}, while classifier-free guidance (CFG) strengthens the conditional score during sampling~\cite{ho2022classifier}. 
Despite being trained to approximate a data distribution, these models can exhibit memorization, reproducing training images or generating highly similar outputs in response to certain prompts~\cite{carlini2023extracting,somepalli2023diffusion,webster2023reproducible}.

Memorization can differ in its spatial extent, leading to global and local memorization. Global memorization reproduces a training image at the image level, preserving its overall layout, object arrangement, and often its fine details. Local memorization reproduces only a spatially confined region, while the remaining parts of the generated image can vary across samples~\cite{chen2025exploring,kim2026localizing}. The copied region may preserve near-verbatim details or retain a distinctive composition while changing colors, textures, and other surface attributes, corresponding to the template-like behavior reported in prior works~\cite{webster2023reproducible,ren2024unveiling,zhao2026you}. Because local memorization can be hidden within an otherwise novel image, identifying its spatial extent is essential for detecting copied content and applying mitigation selectively.

Several factors make memorization easier to learn and retrieve. Repeated training examples receive greater effective weight during optimization, and specific captions can strengthen the association between an image and the words that retrieve it~\cite{carlini2023extracting,somepalli2023understanding}. Deduplication can reduce exact copying, but similar compositions and strong image-text associations may still support template memorization~\cite{webster2023reproducible,han2026adjusting}. These effects are also visible in the denoising network: memorized prompts often yield large conditional-unconditional score differences~\cite{wen2024detecting}, while cross-attention can concentrate on a small number of prompt tokens associated with the copied content~\cite{ren2024unveiling}. Memorization is further shaped by the sampling trajectory. \cite{jain2025classifier} shows that CFG can create an attraction basin that steers a trajectory toward a memorized solution. Delaying or reversing guidance can help the trajectory leave this basin. \cite{han2026adjusting} therefore treats the initial noise as an inference-time control variable and updates it using batch-wise finite differences or per-sample optimization. 

A complementary line of works explains memorization through the geometry of the learned distribution. The Manifold Memorization Hypothesis ~\cite{ross2025geometric} associates memorized samples with reduced local intrinsic dimensionality; ~\cite{jeon2024understanding} relates them to sharp conditional probability landscapes. \cite{kim2026localizing} extends this perspective to spatial localization by estimating coordinate-wise curvature differences. These studies establish a progression from observing copied outputs to measuring conditioning-dependent, geometric, and spatial evidence. 

\subsection{Memorization Localization and Mitigation}

Memorization detection methods typically ask whether a generated image contains memorized content. Retrieval-based methods compare the generated image with a training corpus using pixel- or feature-level similarity~\cite{carlini2023extracting,somepalli2023diffusion}; self-supervised descriptors such as SSCD improve robustness to moderate transformations~\cite{pizzi2022self}. Retrieval provides direct evidence of exact or near-exact copying, but requires access to candidate training images and may miss local or template memorization when the copied region occupies only a small part of the output. Retrieval-free detectors instead use model responses, including score differences~\cite{wen2024detecting}, local intrinsic dimensionality~\cite{ross2025geometric}, and probability-landscape sharpness~\cite{jeon2024understanding}. These signals are easier to obtain during sampling, but they generally produce image-level scores and do not indicate where the memorized content appears.

Localization addresses this complementary question by identifying the spatial regions associated with memorization. ~\cite{ren2024unveiling} associates memorization with concentrated attention on trigger or summary tokens, while Bright Ending~\cite{chen2025exploring} uses late-stage cross-attention to construct a spatial map. However, attention may also reflect ordinary semantic alignment and can vary across layers, tokens, and denoising states. \cite{kim2026localizing} instead estimates reference-subtracted coordinate-wise curvature, providing a geometric localization signal at a selected denoising state. TACO follows this coordinate-wise perspective and aggregates corresponding curvature responses across neighboring late-stage states, reducing transient peaks and producing more spatially consistent memorization maps.

Mitigation methods operate during training, after training, or at inference time. Training-time deduplication, data curation, and regularization reduce the chance of learning easily extractable copies~\cite{somepalli2023understanding,wen2024detecting,ren2024unveiling}. Post-training editing and unlearning modify the model itself: concept-erasure and gradient-based methods suppress specified concepts~\cite{gandikota2023erasing,fan2024salun,park2024direct}, example-level unlearning targets selected training data~\cite{alberti2025data}, and representation-localization methods identify memorization subspaces, responsible neurons, or foreground-background mechanisms~\cite{chavhan2024memorized,hintersdorf2024finding,di2026demystifying}. These approaches can produce persistent changes, but generally require target information and may alter unrelated capabilities. Inference-time methods preserve the released model and modify the prompt, guidance, initial noise, or attention pathway. Prompt perturbation and embedding optimization weaken associations between memorized images and their textual triggers~\cite{somepalli2023understanding,wen2024detecting}. Guidance-based methods delay CFG, set it to zero, or apply opposite guidance to help trajectories escape memorization-attracting regions~\cite{jain2025classifier}. Han et al.~\cite{han2026adjusting} modify the initial noise through finite-difference or gradient-based updates. These methods control the sampling trajectory globally, so stronger suppression may also affect prompt-relevant regions. \cite{zhao2026you} introduces a more selective attention intervention. The detector finds token-wise attention spikes dynamically at each denoising step, attenuates the corresponding cross-attention blocks, and constructs an alternative conditional noise prediction. 

TACO provides a spatial control interface that connects detection with inference-time mitigation. It aggregates coordinate-wise curvature evidence across neighboring late-stage states to form a continuous memorization map. The map is thresholded for localization and retained as a soft spatial weight for steering. In the experiments, TACO steers the initial-noise methods of ~\cite{han2026adjusting} and the zero- or opposite-guidance strategies of ~\cite{jain2025classifier}. Thus, TACO reuses existing mitigation objectives while directing their adjustable components toward regions supported by persistent spatial evidence.

\section{Method Details}
\label{SecB}
\subsection{Algorithm Pseudocode}

Algo.~\ref{alg:taco_detection} summarizes the construction of the continuous TACO memorization map. Given a prompt, TACO follows the standard denoising trajectory and estimates prompt-dependent curvature only at the selected late-stage observation states. The resulting curvature responses are processed and spatially refined before being aggregated into a single memorization map.

\begin{algorithm}[t]
\caption{\textsc{TACOLocalization}}
\label{alg:taco_detection}
\begin{algorithmic}[1]
\Require Text prompt $p$, observation states $\mathcal{T}_{\mathrm{obs}}$
\Ensure Continuous memorization map $M$

\State Sample initial latent $x_0$ and initialize $\mathcal{C}\gets\emptyset$
\For{$t=0,\ldots,T-1$}
    \State Evaluate $\epsilon_t^{\mathrm{u}}(x_t)$ and $\epsilon_t^{\mathrm{c}}(x_t,p)$
    \If{$t\in\mathcal{T}_{\mathrm{obs}}$}
        \State $r_t\gets\epsilon_t^{\mathrm{c}}(x_t,p)-\epsilon_t^{\mathrm{u}}(x_t)$
        \State $\widehat{H}_t\gets
        \Call{DiagonalCurvature}{x_t,r_t,\sigma_t}$
        \Comment{Estimate diagonal curvature}
        \State $H_t^{-}(x,c)\gets\operatorname{ReLU}(-\widehat{H}_t(x,c))$
        \State $C_t(x)\gets\frac{1}{C}\sum_{c=1}^{C}H_t^{-}(x,c)$
        \Comment{Aggregate negative curvature across channels}
        \State $\mu_t^C,s_t^C\gets\Call{SpatialStatistics}{C_t}$
        \State $C_t(x)\gets C_t(x)\operatorname{sigmoid}\left(
        \beta\frac{C_t(x)-\mu_t^C}{s_t^C+\epsilon}\right)$
        \Comment{Emphasize salient responses}
        \State $\widetilde{C}_t\gets
        \Call{LatentGuidedRefinement}{C_t,x_t}$
        \State $\mathcal{C}\gets\mathcal{C}\cup\{(t,\widetilde{C}_t)\}$
    \EndIf
    \State $\epsilon_t^{\mathrm{g}}\gets
    \epsilon_t^{\mathrm{u}}+
    s(\epsilon_t^{\mathrm{c}}-\epsilon_t^{\mathrm{u}})$
    \State $x_{t+1}\gets\mathcal{D}_t(x_t,\epsilon_t^{\mathrm{g}})$
\EndFor
\State Sort $\mathcal{T}_{\mathrm{obs}}=\{t_1,\ldots,t_n\}$ in ascending order
\For{$i=1,\ldots,n$}
    \State $w_i\gets 2-\frac{i-1}{n-1}$
\EndFor
\State $M\gets
\frac{\sum_{i=1}^{n}w_i\widetilde{C}_{t_i}}
{\sum_{i=1}^{n}w_i}$
\State \Return $M$
\end{algorithmic}
\end{algorithm}

Algo.~\ref{alg:taco_mitigation} summarizes the TACO steering mechanism for memorization mitigation. Given a continuous memorization map and an intervention produced by an arbitrary baseline method, when the baseline objective gradient is available, TACO preserves the objective-aligned intervention component and spatially redistributes only its orthogonal component according to the memorization map. Otherwise, TACO combines localized attenuation of classifier-free guidance with a transverse correction orthogonal to the prompt-dependent score difference. In both cases, TACO concentrates the intervention on memorization-related regions while preserving the baseline method's original optimization role.

\begin{algorithm}[t]
\caption{\textsc{TACOSteering}}
\label{alg:taco_mitigation}
\begin{algorithmic}[1]
\Require Text prompt $p$, baseline intervention $\Delta_t^{\mathrm{base}}$, current latent $x_t$
\Ensure TACO-steered intervention $\Delta_t^{\mathrm{TACO}}$

\State $M \gets \Call{TACOLocalization}{p}$
\State $\widetilde{M}_t \gets
\Call{PreprocessMap}{M,x_t}$
\Comment{Preprocessing map}

\If{objective gradient $g_t$ is available}
    \State $\Delta_t^{\parallel} \gets
    \frac{\langle\Delta_t^{\mathrm{base}},g_t\rangle}
    {\|g_t\|_2^2+\epsilon}g_t, \quad\Delta_t^{\perp} \gets
    \Delta_t^{\mathrm{base}}-\Delta_t^{\parallel}$
    \State $\widehat{\Delta}_t^{\perp} \gets
    \widetilde{M}_t\odot\Delta_t^{\perp}$
    \If{$\|\widehat{\Delta}_t^{\perp}\|_2>\epsilon$}
        \State $\Delta_t^{\perp,\mathrm{TACO}} \gets
        \frac{\|\Delta_t^{\perp}\|_2}
        {\|\widehat{\Delta}_t^{\perp}\|_2+\epsilon}
        \widehat{\Delta}_t^{\perp}$
        \State $\Delta_t^{\mathrm{TACO}} \gets
        \Delta_t^{\parallel}+\Delta_t^{\perp,\mathrm{TACO}}$
    \Else
        \State $\Delta_t^{\mathrm{TACO}} \gets
        \Delta_t^{\mathrm{base}}$
    \EndIf
\Else
\State $r_t \gets
\epsilon_t^{\mathrm{c}}(x_t,p)-\epsilon_t^{\mathrm{u}}(x_t)$
\State $\mathcal{A}_t \gets
\operatorname{MaskedRMS}(r_t,\widetilde{M}_t)$
\Comment{Measure the map-weighted guidance strength}
\State $q_t \gets \nabla_{x_t}\mathcal{A}_t$
\State $q_t^{\parallel} \gets
\frac{\langle q_t,r_t\rangle}
{\|r_t\|_2^2+\epsilon}r_t,
\quad
q_t^{\perp} \gets q_t-q_t^{\parallel}$
\State $\Delta_t^{\mathrm{local}} \gets
-\lambda_{\mathrm{cfg}}\widetilde{M}_t\odot r_t$
\Comment{Locally attenuate classifier-free guidance}
\State $\Delta_t^{\mathrm{trans}} \gets
\sigma_t\mathcal{A}_t
\frac{q_t^{\perp}}
{\operatorname{RMS}(q_t)+\epsilon}$
\Comment{Apply the normalized transverse correction}
\State $\Delta_t^{\mathrm{TACO}} \gets
\Delta_t^{\mathrm{base}}
+\Delta_t^{\mathrm{local}}
+\Delta_t^{\mathrm{trans}}$
\EndIf

\State \Return $\Delta_t^{\mathrm{TACO}}$
\end{algorithmic}
\end{algorithm}

\subsection{Computational Overhead}

Let $T$ denote the number of sampling steps, $n=|\mathcal{T}_{\mathrm{obs}}|$ the number of observed states, $J$ the number of curvature probes, and $D=C\times H\times W$ the latent dimensionality. Let $\mathcal{F}$ and $\mathcal{V}$ denote the costs of one U-Net forward evaluation and one vector-Jacobian product, respectively. Standard sampling requires $\mathcal{O}(T\mathcal{F})$ computation. TACO localization additionally performs forward evaluations at the observed states, probe-based curvature estimation, and latent-guided refinement over $k\times k$ neighborhoods, yielding an additional cost of
\begin{equation}
\mathcal{O}
\left(
n\mathcal{F}+nJ\mathcal{V}+nDk^2
\right).
\end{equation}

In our implementation, $n=2$, $J=16$, and $k=3$. Curvature estimation therefore requires only $32$ structured vector-Jacobian products, while the remaining map-processing operations are inexpensive relative to U-Net evaluation. The probes are evaluated sequentially, so the accumulated curvature map requires $\mathcal{O}(D)$ additional storage, apart from the autograd graph used by the current observation state.

TACO steering is substantially lighter. Let $D=C\times H\times W$ denote the latent dimensionality and let $L$ denote the number of intervention steps. Apart from the map construction process, when the objective gradient is available, TACO steering consists of element-wise spatial reweighting, inner products, and norm rescaling. It therefore incurs $\mathcal{O}(D)$ additional computation at each intervention step, or $\mathcal{O}(LD)$ over the full sampling process. When the objective gradient is unavailable, the transverse correction additionally requires one gradient evaluation of the map-weighted attraction objective at each applicable intervention step. Apart from this gradient computation, all steering operations remain linear in the latent dimensionality. Thus, TACO can be integrated with different mitigation objectives without altering their optimization procedure or introducing substantial additional memory overhead.

Tab.~\ref{tab:runtime_overhead} reports the average wall-clock generation time per image. In our mitigation evaluation, each prompt produces $10$ samples using seeds $0$ through $9$. The TACO memorization map is computed only once from the generation with seed $0$ and is subsequently reused for the remaining samples of the same prompt. Thus, the curvature-based localization cost is shared across all ten generations, whereas the reported additional runtime mainly reflects the lightweight steering operations.

Across the four mitigation baselines, TACO introduces an additional $0.24$-$0.36$ seconds per image on SD v1.4 and $0.23$-$0.33$ seconds on SD v2.0. The overhead remains consistent despite the different computational profiles of the baseline methods, indicating that TACO steering acts as an efficient plug-and-play module. The prompt-level reuse of the TACO map further reduces the effective cost of spatial localization during multi-sample mitigation.

\begin{table}[t]
\small
\centering
\caption{Average generation time per image in seconds. Values in parentheses denote the additional runtime introduced by TACO calculation steering.}
\label{tab:runtime_overhead}
\begin{tabular}{l l ccc}

\toprule
\multirow{2}{*}{Method} & \multicolumn{2}{c}{SD v1.4} & \multicolumn{2}{c}{SD v2.0} \\
\cmidrule(lr){2-3} \cmidrule(lr){4-5}
& Base & +Ours & Base & +Ours \\
\midrule
No Mitigation  & $0.97$ & - & $0.96$ & - \\ \midrule
\texttt{Init Batch }  & $1.03$ & $1.27_{(+0.24)}$ & $0.97$ & $1.20_{(+0.23)}$ \\
\texttt{Init Sample }   & $2.04$ & $2.40_{(+0.36)}$ & $1.79$ & $2.05_{(+0.26)}$ \\
\texttt{Basin Zero}         & $0.95$ & $1.26_{(+0.31)}$ & $0.90$ & $1.23_{(+0.33)}$ \\
\texttt{Basin Neg }      & $0.95$ & $1.26_{(+0.31)}$ & $0.90$ & $1.20_{(+0.30)}$ \\

\bottomrule
\end{tabular}%
\label{supp_result1}
\end{table}

\section{Experimental Details and Additional Results}
\label{SecC}

\subsection{Experimental Setup}

\paragraph{Models and Runtime Environment.}
Following the corresponding benchmark settings, we evaluate memorization localization using SD v1.4
(\texttt{CompVis/stable-diffusion-v1-4}) and SD v2.1
(\texttt{Manojb/stable-diffusion-2-1-base}). For memorization mitigation, we use SD v1.4 and SD v2.0 (\texttt{Manojb/stable-diffusion-2-base}). In both localization and mitigation settings, we use a DDIM scheduler with 50 inference steps and a CFG guidance scale of $7.5$. Sampling steps are indexed according to their execution order from $0$ to $49$. All experiments are implemented in PyTorch  2.4.0 \cite{NEURIPS2019_9015} using the Hugging Face Diffusers library (version 0.36.0) \cite{von-platen-etal-2022-diffusers} and run under Ubuntu 22.04.5 LTS on a server equipped with one single NVIDIA GeForce RTX 4090 GPU with 24GB VRAM.

\paragraph{TACO for Memorization Localization.} 
We evaluate memorization localization on the benchmark provided by~\cite{kim2026localizing}. Each benchmark instance consists of a prompt--seed pair and its corresponding ground-truth binary mask, where a value of one indicates a memorized region and zero indicates a non-memorized region. For each prompt, we evaluate at most four generation seeds provided by the benchmark. The benchmark is evaluated under two settings: \textit{TV only}, which contains samples with localized memorized regions, and \textit{All}, which additionally includes non-memorized samples (For non-memorized samples, The generation seed is fixed at $0$). The latter setting evaluates whether a method can distinguish memorized content from ordinary generated regions.

We use mean Intersection over Union (mIoU) and pixel-level accuracy (Acc) to quantify memorization localization.
Given a predicted continuous memorization map $M_i$ and its ground-truth binary mask $Y_i$ for sample $i$, we obtain a binary prediction mask using a threshold $\tau$:
\begin{equation}
\widehat{Y}_{i,\tau}(x)
=
\mathbb{I}\left[M_i(x)>\tau\right],
\end{equation}
where $Y_i(x)\in\{0,1\}$ indicates whether spatial location $x$ belongs to a memorized region. For each sample, the Intersection over Union (IoU) is defined as $
\operatorname{IoU}_i
=
\frac{
|\widehat{Y}_i\cap Y_i|
}{
|\widehat{Y}_i\cup Y_i|
}$. The mean IoU averages the per-sample IoU over the $N$ evaluation samples: $
\operatorname{mIoU}
=
\frac{1}{N}
\sum_{i=1}^{N}\operatorname{IoU}_i
$. Pixel-level accuracy measures the fraction of correctly classified pixels across all samples as $
\operatorname{Acc}
=
\frac{
\sum_{i=1}^{N}\sum_{x}
\mathbb{I}\left[\widehat{Y}_i(x)=Y_i(x)\right]
}{
\sum_{i=1}^{N}\sum_x 1
}$.

For each metric, the threshold $\tau$ is selected independently to optimize the corresponding metric on its evaluation set. We report both mIoU and Acc to assess complementary aspects of localization quality: mIoU emphasizes spatial overlap, whereas Acc measures overall pixel-wise classification correctness.

\begin{table*}[t]
\caption{Quantitative evaluation of Pareto fronts in memorization mitigation.}
\label{tab:pareto}
\centering
\small
\begin{tabular}{l l *{4}{>
{\centering\arraybackslash}m{1.8cm}}}
\toprule
\multirow{2}{*}{Model} & \multirow{2}{*}{Method Pair ($\mathrm{ours}-\mathrm{base}$)} & \multicolumn{2}{c}{\textbf{SSCD-CLIP}} & \multicolumn{2}{c}{\textbf{SSCD-ImageReward}} \\
\cmidrule(lr){3-4} \cmidrule(lr){5-6}
& & NPHV($\uparrow$) & PDA($\uparrow$) & NPHV($\uparrow$) & PDA($\uparrow$) \\
\midrule
\multirow{4}{*}{SD v1.4}
& \texttt{Init Batch }  & $0.061$ & $0.714$ & $0.145$ & $0.714$ \\
& \texttt{Init Sample }   & $0.136$ & $1.000$ & $0.181$ & $1.000$ \\
& \texttt{Basin Zero}         & $0.065$ & $0.285$ & $0.091$ & $0.714$ \\
& \texttt{Basin Neg }      & $0.042$ & $0.857$ & $0.061$ & $0.857$ \\
\midrule
\multirow{4}{*}{SD v2.0}
& \texttt{Init Batch }  & $0.274$ & $1.000$ & $0.307$ & $1.000$ \\
& \texttt{Init Sample }   & $0.230$ & $1.000$ & $0.135$ & $1.000$ \\
& \texttt{Basin Zero}         & $0.103$ & $0.571$ & $0.117$ & $0.571$ \\

& \texttt{Basin Neg} & $0.063$ & $0.285$ & $0.082$ & $0.285$ \\
\bottomrule
\end{tabular}%
\label{supp_result1}
\end{table*}

\paragraph{TACO steering for Memorization Mitigation.} We use the benchmark provided by~\cite{webster2023reproducible}, which contains text prompts associated with both local and global memorization. For each prompt, we generate $10$ samples using seeds $0$ through $9$, and compare each mitigated sample with the reference image. To avoid repeated computation, we construct the TACO map once from the sampling trajectory generated with seed $0$. The same map is then reused to steer mitigation for all ten seeds of that prompt.

We integrate TACO steering into several existing mitigation methods to evaluate its effectiveness across different intervention strategies. Unless otherwise specified, we follow the default configurations of each baseline. 

To characterize the trade-off between memorization suppression and generation quality, we vary the mitigation strength of each baseline and evaluate each configuration both with and without TACO steering: 
\begin{itemize}
    \item For \texttt{Init Batch}, we vary the intervention step over $\{0,1,2,3,4,5,6\}$, with $4$ as the default; earlier interventions generally favor quality preservation.
    \item  For \texttt{Init Sample}, we vary the learning rate for initial-noise optimization and the early-stopping target loss, denoted as the $(\texttt{lr}, \texttt{target loss})$ pair. We evaluate $\{(0.006,1.2), (0.006,0.9), (0.008,0.9), (0.010,0.9),\\ (0.012,0.9), (0.014,0.9), (0.014,0.6)\}$, with $(0.01,0.9)$ as the default. Lower learning rates and higher target losses generally favor quality preservation.
    \item For \texttt{Basin Zero} and \texttt{Basin Neg}, we vary the intervention duration in the diffusion timestep scale over $\{50,100,150,200,250,300,350\}$ and $\{50,75,85,100,125,150,175,200\}$, respectively, with $100$ as the default for both methods. An intervention duration of $d$ corresponds to intervening over the initial $d$ diffusion timesteps (e.g., from timestep $1000$ to $1000-d$). Shorter durations generally better preserve image quality.
\end{itemize}  
We compare the resulting Pareto frontiers in the SSCD-CLIP and SSCD-ImageReward spaces to assess whether TACO steering improves memorization suppression while preserving semantic alignment and generation quality. 
For SSCD, we use the pretrained checkpoint \texttt{sscd\_disc\_large.torchscript.pt} released by~\cite{pizzi2022self}, which extracts image descriptors for measuring visual similarity between each mitigated image and its corresponding unmitigated reference. Lower SSCD similarity indicates stronger suppression of memorized content. For semantic alignment, we use the \texttt{ViT-B-32} model pretrained on \texttt{laion2b\_s34b\_b79k} to measure image-text similarity between the generated image and its input prompt. For perceptual preference quality, we use the pretrained \texttt{ImageReward-v1.0} model, which predicts human preference for text-to-image generations. Higher CLIP and ImageReward scores indicate better semantic alignment and perceptual preference, respectively. 


To quantitatively compare the resulting Pareto frontiers, we report the normalized Pareto hypervolume difference (NPHV) based on the hypervolume indicator~\cite{zitzler2007hypervolume}:
\begin{equation}
\operatorname{NPHV}
=
\frac{
HV_{\mathrm{ours}}-HV_{\mathrm{base}}
}{
HV_{\mathrm{ours}}+HV_{\mathrm{base}}
},
\end{equation}
where $HV_{\mathrm{ours}}$ and $HV_{\mathrm{base}}$ denote the hypervolumes of the TACO-steered mitigation method and its corresponding original baseline, respectively. For each objective pair, we jointly apply min-max normalization to both methods using shared minimum and maximum values; the objective to be minimized (SSCD) are reversed as $1-x$, while objectives to be maximized (CLIP and ImageReward) retain their normalized values. The hypervolumes are computed using a common reference point $(-0.01,-0.01)$, slightly below the normalized objective space, and measure the area covered by the rectangles between this point and the non-dominated solutions. A positive NPHV indicates that the proposed method achieves a larger hypervolume than the baseline, while a negative value indicates the opposite.

We additionally report the Pareto dominance advantage (PDA), which measures the fraction of the proposed method's operating points that dominate at least one baseline operating point:
\begin{equation}
\operatorname{PDA}
=
\frac{
\sum_{p\in\mathcal{P}_{\mathrm{ours}}}
\mathbb{I}
\left[
\exists q\in\mathcal{P}_{\mathrm{base}}:
p\succ q
\right]
}{
|\mathcal{P}_{\mathrm{ours}}|
},
\end{equation}
where $\mathcal{P}_{\mathrm{ours}}$ and $\mathcal{P}_{\mathrm{base}}$ denote the respective sets of operating points, and $p\succ q$ indicates that $p$ achieves no worse SSCD and no worse quality score than $q$, with at least one strict improvement. A higher PDA indicates that a larger proportion of the proposed method's operating points dominate at least one baseline operating point. The corresponding results are shown in Tab. \ref{supp_result1}.

\subsection{Additional Results}

Tab.~\ref{tab:pareto} summarizes the Pareto-front comparison between each baseline and its TACO-steered counterpart. All eight comparisons yield positive NPHV values in both evaluation spaces, indicating that TACO consistently enlarges the trade-off region between memorization suppression and generation quality. For SD v1.4, the largest improvements are obtained with \texttt{Init Sample}, whose NPHV reaches $0.136$ in SSCD-CLIP and $0.181$ in SSCD-ImageReward. Its PDA is $1.000$ in both spaces, meaning that every TACO-steered operating point dominates at least one point of the corresponding baseline frontier. The other three baselines also benefit from TACO. These results show that the spatial map remains useful across both initial-noise and guidance-based interventions.

The improvements are even larger for SD v2.0. TACO-steered \texttt{Init Batch} obtains the highest NPHV, with gains of $0.274$ in SSCD-CLIP and $0.307$ in SSCD-ImageReward. \texttt{Init Sample} also achieves substantial improvements of $0.230$ and $0.135$, respectively, with PDA equal to $1.000$ in both spaces. The two guidance-based methods remain consistently improved. Thus, the benefit of TACO is observed across different mitigation mechanisms and models, rather than being tied to a particular baseline.

Overall, the uniformly positive NPHV and PDA values demonstrate that TACO improves the suppression-quality trade-off across all evaluated settings. The results support the effectiveness of using the trajectory-aggregated spatial memorization map to steer existing mitigation objectives toward the regions most strongly associated with memorized content.

Tab.~\ref{pareto1}-\ref{pareto4} provide the detailed SSCD, CLIP, and ImageReward scores for the four groups respectively. These tables report the operating points used to construct the Pareto frontiers and provide the underlying quantitative results for the comparisons summarized in Tab.~\ref{tab:pareto}.

\subsection{Additional Visualizations}

\begin{figure*}[ht]
     \centering
     \includegraphics[width=\textwidth]{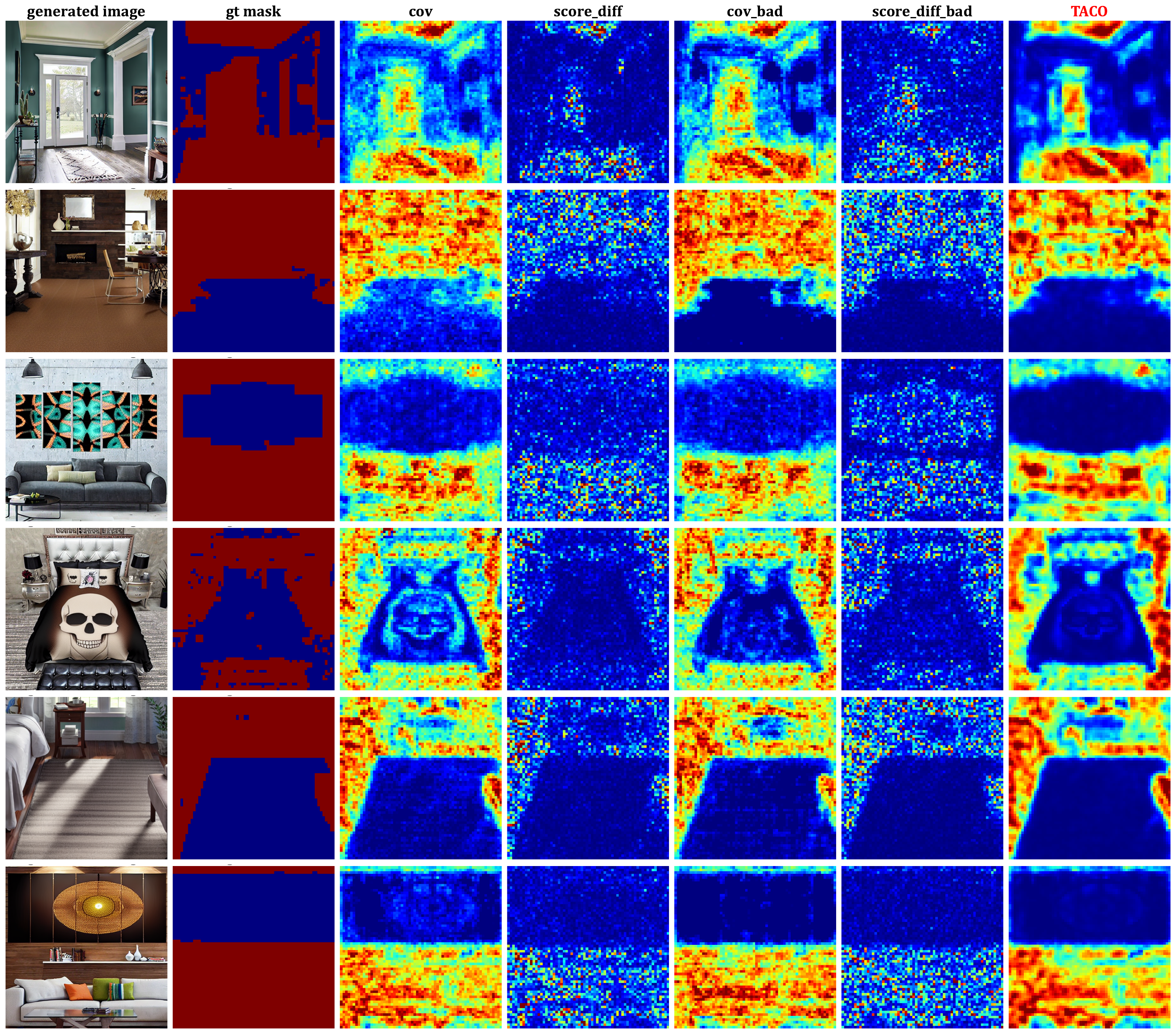}
     \caption{Additional qualitative comparison of memorization localization maps produced by different methods.}
     \label{fig:additional_detection}
\end{figure*}

\begin{figure*}[ht]
     \centering
     \includegraphics[width=\textwidth]{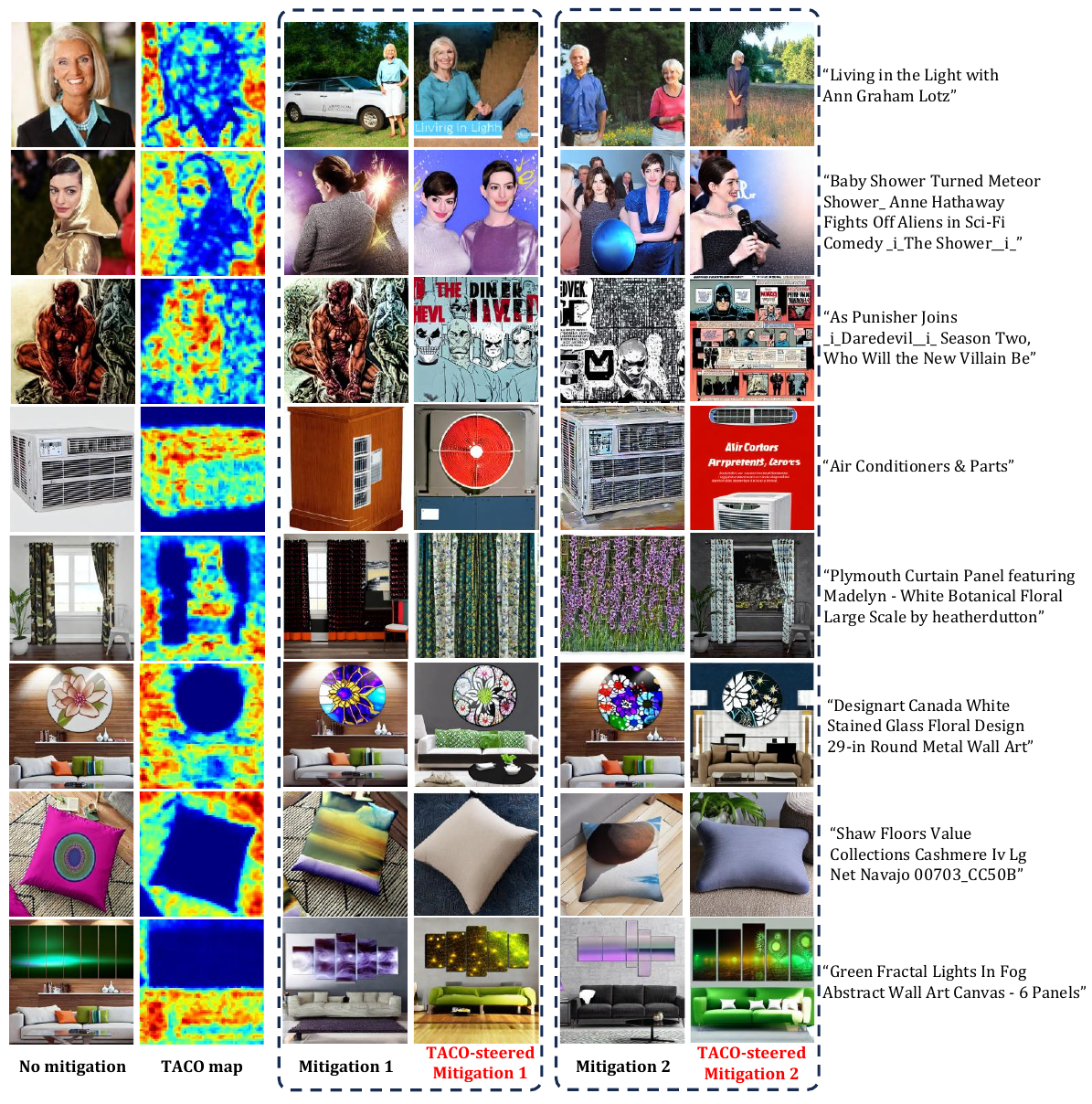}
     \caption{Additional qualitative comparison of existing memorization mitigation methods with and without TACO steering. Each pair compares the original mitigation baseline with its TACO-steered counterpart using the same generation seed.}
     \label{fig:additional_mitigation}
\end{figure*}

Fig.~\ref{fig:additional_detection} compares the memorization maps produced by TACO and other localization methods. TACO generates spatially coherent responses with clearer boundaries around the memorized regions, while reducing activations in surrounding non-memorized areas. Fig.~\ref{fig:additional_mitigation} shows the final generated images obtained with and without TACO steering for several existing mitigation methods. The TACO-steered results reduce the copied content highlighted by the map while preserving the surrounding image structure. Together, these examples demonstrate that TACO provides both interpretable local memorization maps and an effective spatial control signal for guiding existing mitigation methods.

\label{SecD}

\begin{table*}[t]
\centering

\small
\caption{Detailed statistics of Pareto fronts comparison between the original \texttt{Init Batch} and TACO-steered \texttt{Init Batch}.}
\label{pareto1}
\resizebox{0.9\textwidth}{!}{
\begin{tabular}{lccccccc}
\toprule
\multirow{2}{*}{Method} & \multirow{2}{*}{Param} & \multicolumn{3}{c}{SD v1.4} & \multicolumn{3}{c}{SD v2.0} \\
\cmidrule(lr){3-5} \cmidrule(lr){6-8}
& & SSCD($\downarrow$) & CLIP($\uparrow$) & ImageReward($\uparrow$) & SSCD($\downarrow$) & CLIP($\uparrow$) & ImageReward($\uparrow$) \\
\midrule
No mitigation & - & $0.5106$ & $0.2999$ & $0.4548$ & $0.3196$ & $0.3039$ & $0.1406$ \\
\midrule
\multirow{7}{*}{\texttt{Init Batch}}
& $0$ & $0.3807$ & $0.2876$ & $-0.1419$ & $0.2899$ & $0.2982$ & $-0.0033$ \\
& $1$ & $0.3441$ & $0.2812$ & $-0.2050$ & $0.2809$ & $0.2954$ & $-0.0528$ \\
& $2$ & $0.3061$ & $0.2743$ & $-0.2809$ & $0.2688$ & $0.2919$ & $-0.1260$ \\
& $3$ & $0.2700$ & $0.2670$ & $-0.3632$ & $0.2534$ & $0.2865$ & $-0.2109$ \\
& $4$ & $0.2376$ & $0.2604$ & $-0.4377$ & $0.2344$ & $0.2824$ & $-0.2839$ \\
& $5$ & $0.2050$ & $0.2522$ & $-0.5372$ & $0.2153$ & $0.2771$ & $-0.3662$ \\
& $6$ & $0.1777$ & $0.2452$ & $-0.6242$ & $0.1935$ & $0.2714$ & $-0.4606$ \\
\midrule
\multirow{7}{*}{\texttt{Init Batch+Ours}}
& $0$ & $0.3473$ & $0.2821$ & $-0.1592$ & $0.2802$ & $0.2991$ & $0.0363$ \\
& $1$ & $0.3052$ & $0.2754$ & $-0.2216$ & $0.2681$ & $0.2970$ & $-0.0144$ \\
& $2$ & $0.2676$ & $0.2700$ & $-0.2820$ & $0.2538$ & $0.2944$ & $-0.0584$ \\
& $3$ & $0.2340$ & $0.2633$ & $-0.3604$ & $0.2360$ & $0.2903$ & $-0.1280$ \\
& $4$ & $0.2050$ & $0.2568$ & $-0.4346$ & $0.2174$ & $0.2862$ & $-0.1905$ \\
& $5$ & $0.1768$ & $0.2499$ & $-0.5185$ & $0.1988$ & $0.2823$ & $-0.2666$ \\
& $6$ & $0.1553$ & $0.2436$ & $-0.5974$ & $0.1803$ & $0.2768$ & $-0.3613$ \\
\bottomrule
\end{tabular}
}
\end{table*}

\begin{table*}[t]
\centering

\small
\caption{Detailed statistics of Pareto fronts comparison between the original \texttt{Init Sample} and TACO-steered \texttt{Init Sample}.}
\resizebox{0.9\textwidth}{!}{
\label{pareto2}
\begin{tabular}{lccccccc}
\toprule
\multirow{2}{*}{Method} & \multirow{2}{*}{Param} & \multicolumn{3}{c}{SD v1.4} & \multicolumn{3}{c}{SD v2.0} \\
\cmidrule(lr){3-5} \cmidrule(lr){6-8}
& & SSCD($\downarrow$) & CLIP($\uparrow$) & ImageReward($\uparrow$) & SSCD($\downarrow$) & CLIP($\uparrow$) & ImageReward($\uparrow$) \\
\midrule
No mitigation & - & $0.5106$ & $0.2999$ & $-0.0060$ & $0.3196$ & $0.3039$ & $0.1406$ \\
\midrule
\multirow{7}{*}{\texttt{Init Sample}}
& $(0.006, 1.2)$ & $0.2903$ & $0.2807$ & $-0.1829$ & $0.2593$ & $0.2981$ & $0.0628$ \\
& $(0.006, 0.9)$ & $0.2211$ & $0.2697$ & $-0.3081$ & $0.2283$ & $0.2887$ & $-0.0763$ \\
& $(0.008, 0.9)$ & $0.2085$ & $0.2673$ & $-0.3253$ & $0.2172$ & $0.2867$ & $-0.1111$ \\
& $(0.010, 0.9)$ & $0.1982$ & $0.2653$ & $-0.3510$ & $0.2091$ & $0.2859$ & $-0.1181$ \\
& $(0.012, 0.9)$ & $0.1845$ & $0.2618$ & $-0.3894$ & $0.1971$ & $0.2844$ & $-0.1447$ \\
& $(0.014, 0.9)$ & $0.1705$ & $0.2592$ & $-0.4237$ & $0.1862$ & $0.2828$ & $-0.1870$ \\
& $(0.014, 0.6)$ & $0.0987$ & $0.2328$ & $-0.7871$ & $0.1275$ & $0.2634$ & $-0.5711$ \\
\midrule
\multirow{7}{*}{\texttt{Init Sample+Ours}}
& $(0.006, 1.2)$ & $0.2862$ & $0.2822$ & $-0.1471$ & $0.2566$ & $0.3001$ & $0.0804$ \\
& $(0.006, 0.9)$ & $0.2063$ & $0.2719$ & $-0.2436$ & $0.2103$ & $0.2936$ & $-0.0522$ \\
& $(0.008, 0.9)$ & $0.1893$ & $0.2691$ & $-0.2718$ & $0.2012$ & $0.2923$ & $-0.0843$ \\
& $(0.010, 0.9)$ & $0.1780$ & $0.2680$ & $-0.2764$ & $0.1903$ & $0.2905$ & $-0.0998$ \\
& $(0.012, 0.9)$ & $0.1660$ & $0.2650$ & $-0.3045$ & $0.1797$ & $0.2898$ & $-0.1253$ \\
& $(0.014, 0.9)$ & $0.1579$ & $0.2635$ & $-0.3312$ & $0.1713$ & $0.2884$ & $-0.1426$ \\
& $(0.014, 0.6)$ & $0.0966$ & $0.2403$ & $-0.6422$ & $0.1190$ & $0.2674$ & $-0.5581$ \\
\bottomrule
\end{tabular}
}
\end{table*}

\begin{table*}[t]
\centering
\small

\caption{Detailed statistics of Pareto fronts comparison between the original \texttt{Basin Zero} and TACO-steered \texttt{Basin Zero}.}
\label{pareto3}
\resizebox{0.9\textwidth}{!}{
\begin{tabular}{lccccccc}
\toprule
\multirow{2}{*}{Method} & \multirow{2}{*}{Param} & \multicolumn{3}{c}{SD v1.4} & \multicolumn{3}{c}{SD v2.0} \\
\cmidrule(lr){3-5} \cmidrule(lr){6-8}
& & SSCD($\downarrow$) & CLIP($\uparrow$) & ImageReward($\uparrow$) & SSCD($\downarrow$) & CLIP($\uparrow$) & ImageReward($\uparrow$) \\
\midrule
No mitigation & - & $0.5106$ & $0.2999$ & $0.4548$ & $0.3196$ & $0.3039$ & $0.1406$ \\
\midrule
\multirow{7}{*}{\texttt{Basin Zero}}
& $50$ & $0.4871$ & $0.2984$ & $-0.0259$ & $0.3132$ & $0.3027$ & $0.1097$ \\
& $100$ & $0.4410$ & $0.2926$ & $-0.0801$ & $0.3009$ & $0.3003$ & $0.0411$ \\
& $150$ & $0.3909$ & $0.2856$ & $-0.1654$ & $0.2847$ & $0.2967$ & $-0.0369$ \\
& $200$ & $0.2975$ & $0.2685$ & $-0.3687$ & $0.2431$ & $0.2868$ & $-0.2177$ \\
& $250$ & $0.2352$ & $0.2545$ & $-0.5305$ & $0.1991$ & $0.2766$ & $-0.3889$ \\
& $300$ & $0.1522$ & $0.2294$ & $-0.8368$ & $0.1342$ & $0.2570$ & $-0.7036$ \\
& $350$ & $0.1106$ & $0.2121$ & $-1.0705$ & $0.0981$ & $0.2408$ & $-0.9441$ \\
\midrule
\multirow{7}{*}{\texttt{Basin Zero+Ours}}
& $50$ & $0.4502$ & $0.2965$ & $-0.0242$ & $0.2988$ & $0.3020$ & $0.0970$ \\
& $100$ & $0.3292$ & $0.2821$ & $-0.1628$ & $0.2592$ & $0.2982$ & $0.0204$ \\
& $150$ & $0.2496$ & $0.2681$ & $-0.3099$ & $0.2236$ & $0.2912$ & $-0.1073$ \\
& $200$ & $0.1432$ & $0.2412$ & $-0.6217$ & $0.1560$ & $0.2749$ & $-0.3902$ \\
& $250$ & $0.0972$ & $0.2243$ & $-0.8469$ & $0.1118$ & $0.2607$ & $-0.6316$ \\
& $300$ & $0.0618$ & $0.1990$ & $-1.1869$ & $0.0709$ & $0.2383$ & $-0.9959$ \\
& $350$ & $0.0494$ & $0.1830$ & $-1.4166$ & $0.0592$ & $0.2240$ & $-1.2163$ \\
\bottomrule
\end{tabular}
}
\end{table*}

\begin{table*}[t]
\centering
\small

\caption{Detailed statistics of Pareto fronts comparison between the original \texttt{Basin Neg} and TACO-steered \texttt{Basin Neg}.}
\label{pareto4}
\resizebox{0.9\textwidth}{!}{
\begin{tabular}{lccccccc}
\toprule
\multirow{2}{*}{Method} & \multirow{2}{*}{Param} & \multicolumn{3}{c}{SD v1.4} & \multicolumn{3}{c}{SD v2.0} \\
\cmidrule(lr){3-5} \cmidrule(lr){6-8}
& & SSCD($\downarrow$) & CLIP($\uparrow$) & ImageReward($\uparrow$) & SSCD($\downarrow$) & CLIP($\uparrow$) & ImageReward($\uparrow$) \\
\midrule
No mitigation & - & $0.5106$ & $0.2999$ & $0.4548$ & $0.3196$ & $0.3039$ & $0.1406$ \\
\midrule
\multirow{7}{*}{\texttt{Basin Neg}}
& $50$ & $0.4550$ & $0.2955$ & $-0.0480$ & $0.3043$ & $0.3022$ & $0.0825$ \\
& $75$ & $0.4110$ & $0.2905$ & $-0.1021$ & $0.2921$ & $0.2997$ & $0.0320$ \\
& $85$ & $0.3479$ & $0.2815$ & $-0.1999$ & $0.2721$ & $0.2946$ & $-0.0581$ \\
& $100$ & $0.2801$ & $0.2687$ & $-0.3389$ & $0.2370$ & $0.2866$ & $-0.2039$ \\
& $125$ & $0.2085$ & $0.2534$ & $-0.5139$ & $0.1936$ & $0.2751$ & $-0.3836$ \\
& $150$ & $0.1520$ & $0.2360$ & $-0.7199$ & $0.1482$ & $0.2610$ & $-0.6138$ \\
& $175$ & $0.1098$ & $0.2191$ & $-0.9189$ & $0.1136$ & $0.2466$ & $-0.8275$ \\
\midrule
\multirow{7}{*}{\texttt{Basin Neg+Ours}}
& $50$ & $0.4095$ & $0.2918$ & $-0.0673$ & $0.2886$ & $0.3004$ & $0.0595$ \\
& $75$ & $0.3380$ & $0.2830$ & $-0.1479$ & $0.2661$ & $0.2978$ & $0.0153$ \\
& $85$ & $0.2680$ & $0.2712$ & $-0.2735$ & $0.2397$ & $0.2914$ & $-0.0983$ \\
& $100$ & $0.1964$ & $0.2562$ & $-0.4470$ & $0.1962$ & $0.2806$ & $-0.2621$ \\
& $125$ & $0.1429$ & $0.2392$ & $-0.6580$ & $0.1553$ & $0.2683$ & $-0.4703$ \\
& $150$ & $0.1019$ & $0.2224$ & $-0.8716$ & $0.1172$ & $0.2534$ & $-0.7219$ \\
& $175$ & $0.0771$ & $0.2076$ & $-1.0818$ & $0.0893$ & $0.2391$ & $-0.9440$ \\
\bottomrule

\end{tabular}
}
\end{table*}

\end{appendix}



\end{document}